\documentclass[runningheads]{llncs}

\usepackage{eccv}
\usepackage{marvosym}

\usepackage{eccvabbrv}

\usepackage{graphicx}
\usepackage{booktabs}
\usepackage{enumitem}
\usepackage{multirow}
\usepackage{subcaption}
\usepackage{ulem}
\usepackage{wrapfig}
\usepackage{tabularx} 
\usepackage[accsupp]{axessibility}  

\usepackage{hyperref}

\usepackage{orcidlink}

\begin{document}

\title{ControlNet$++$: Improving Conditional Controls with Efficient Consistency Feedback\\ \textcolor{magenta}{\small Project Page: \href{https://liming-ai.github.io/ControlNet_Plus_Plus/}{liming-ai.github.io/ControlNet\_Plus\_Plus}}\vspace{-0.4cm}}


\titlerunning{ControlNet$++$}
\author{
    Ming Li\inst{1} \and
    Taojiannan Yang\inst{1} \and
    Huafeng Kuang\inst{2} \and
    Jie Wu\inst{2} \and \\
    Zhaoning Wang\inst{1} \and
    Xuefeng Xiao\inst{2} \and
    Chen Chen\inst{1}\orcidlink{0000-0003-3957-7061}
}

\authorrunning{Li et al.}

\institute{Center for Research in Computer Vision, University of Central Florida \and
ByteDance}

\maketitle

\vspace{-0.8cm}

\begin{abstract}

    To enhance the controllability of text-to-image diffusion models, existing efforts like ControlNet incorporated image-based conditional controls.
    In this paper, we reveal that existing methods still face significant challenges in generating images that align with the image conditional controls.
    To this end, we propose ControlNet++, a novel approach that improves controllable generation by explicitly optimizing pixel-level cycle consistency between generated images and conditional controls.
    Specifically, for an input conditional control, we use a pre-trained discriminative reward model to extract the corresponding condition of the generated images, and then optimize the consistency loss between the input conditional control and extracted condition. A straightforward implementation would be generating images from random noises and then calculating the consistency loss, but such an approach requires storing gradients for multiple sampling timesteps, leading to considerable time and memory costs.
    To address this, we introduce an efficient reward strategy that deliberately disturbs the input images by adding noise, and then uses the single-step denoised images for reward fine-tuning. This avoids the extensive costs associated with image sampling, allowing for more efficient reward fine-tuning.
    Extensive experiments show that ControlNet++ significantly improves controllability under various conditional controls. For example, it achieves improvements over ControlNet by 11.1\% mIoU, 13.4\% SSIM, and 7.6\% RMSE, respectively, for segmentation mask, line-art edge, and depth conditions. All the code, models, demo and organized data have been open sourced on our \textcolor{magenta}{\href{https://github.com/liming-ai/ControlNet_Plus_Plus}{Github Repo}}.

\vspace{-0.2cm}
\keywords{Controllable Generation \and Diffusion Model \and ControlNet}
\vspace{-0.2cm}
\end{abstract}

\vspace{-0.5cm}

\begin{figure*}[t]\centering
    \includegraphics[width=0.9\linewidth]{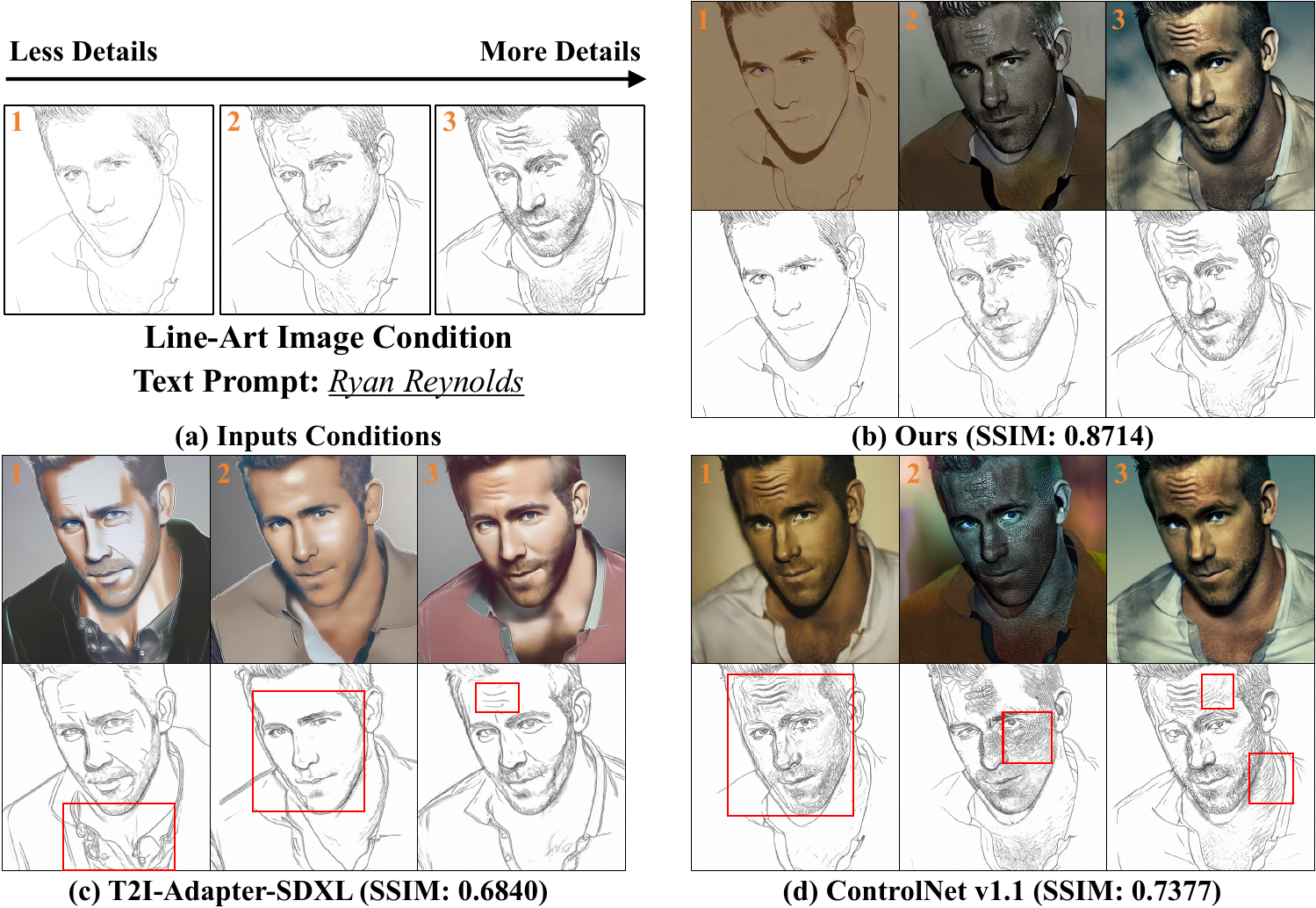}
    \vspace{-0.3cm}
    \caption{
    \textbf{(a)} Given the same input image condition and text prompt, \textbf{(b)} the extracted conditions of our generated images are more consistent with the inputs, \textbf{(c,d)} while other methods fail to achieve accurate controllable generation. 
    SSIM scores measure the similarity between all input edge conditions and the extracted edge conditions. All the line edges are extracted by the same line detection model used by ControlNet~\cite{controlnet}.
    }
    \vspace{-0.7cm}
    \label{fig:teaser}
\end{figure*}

\vspace{-0.2cm}
\section{Introduction}
\label{sec:intro}
\vspace{-0.3cm}
The emergence and improvements of diffusion models~\cite{diffusion,diffusion_beat_gan,sd}, along with the introduction of large-scale image-text datasets~\cite{laion400m,laion5b}, has catalyzed significant strides in text-to-image generation. 
Nonetheless, as the proverb ``an image is worth a thousand words'' conveys, it's challenging to depict an image accurately and in detail through language alone, and this dilemma also perplexes existing text-to-image diffusion models~\cite{sd,imagen}. To this end, many studies focus on incorporating conditional controls such as segmentation mask into text-to-image diffusion models~\cite{controlnet,t2i_adapter,ip_adapter,unicontrol,composer}. Despite the diversity in these methods, the core objective remains to facilitate more accurate and controllable image generation with explicit image-based conditional controls. 

Achieving controllable image generation could involve retraining diffusion models from scratch~\cite{sd,unicontrol}, but this comes with high computational demands and a scarcity of large public datasets~\cite{controlnet}. In light of this, a feasible strategy is fine-tuning pre-trained text-to-image models~\cite{humansd,reco} or introducing trainable modules~\cite{controlnet,t2i_adapter,ip_adapter} like ControlNet~\cite{controlnet}. 
However, despite these studies have explored the feasibility of controllability~\cite{controlnet,t2i_adapter,ip_adapter} in text-to-image diffusion models and expanded various applications~\cite{humansd,unicontrol,composer}, a significant gap remains in achieving precise and fine-grained control. As shown in Fig.~\ref{fig:teaser}, existing methods of controllable generation (\emph{e.g.}, ControlNet~\cite{controlnet} and T2I-Adapter~\cite{t2i_adapter}) still struggle to accurately generate images that are consistent with the input image condition. For example, T2I-Adapter-SDXL consistently produced incorrect wrinkles on the forehead in all generated images, while ControlNet v1.1 introduced many wrong details.
Regrettably, current efforts lack specific methods for improving controllability, which impedes progress in this research area.

To address this issue, we model image-based controllable generation as an image translation task~\cite{image_translation} from input conditional controls to output generated images. 
Inspired by CycleGAN~\cite{cyclegan}, we propose to employ pre-trained discriminative models to extract the condition from the generated images and directly optimize the cycle consistency loss for better controllability.
The idea is that if we translate images from one domain to the other (condition $c_v$ $\rightarrow$ generated image $x'_0$), and back again (generated image $x'_0$ $\rightarrow$ condition $c_v'$) we should arrive where we started ($c'_v=c_v$), as shown in Fig.~\ref{fig:method_high-level}.
For example, given a segmentation mask as a conditional control, we can employ existing methods such as ControlNet~\cite{controlnet} to generate corresponding images. Then the predicted segmentation masks of these generated images can be obtained by a pre-trained segmentation model. Ideally, the predicted segmentation masks and the input segmentation masks should be consistent. Hence, the cycle consistency loss can be formulated as the per-pixel classification loss between the input and predicted segmentation masks. Unlike existing related works~\cite{controlnet,t2i_adapter,gligen,unicontrol,unicontrolnet} that implicitly achieve controllability by introducing conditional controls into the latent-space denoising process, our method \textbf{explicitly} optimizes controllability at the pixel-space for better performance, as demonstrated in Fig.~\ref{fig:comparasion}.

To implement pixel-level loss within the context of diffusion models, an intuitive approach involves executing the diffusion model's inference process, starting from random Gaussian noise and performing multiple sampling steps to obtain the final generated images, following recent works focusing on improving image quality with human feedback~\cite{imagereward,AlignProp,DRaFT}. However, multiple samplings can lead to efficiency issues, and require the storage of gradients at every timestep and thus significant time and GPU memory consumption.
We demonstrate that initiating sampling from random Gaussian noise is unnecessary. Instead, by directly adding noise to training images to disturb their consistency with input conditional controls and then using single-step denoised images to reconstruct the consistency, we can conduct more efficient reward fine-tuning.  Our contributions are summarized as:
\vspace{-0.1cm}

\begin{figure}[t!]\centering
    \includegraphics[width=1.0\linewidth]{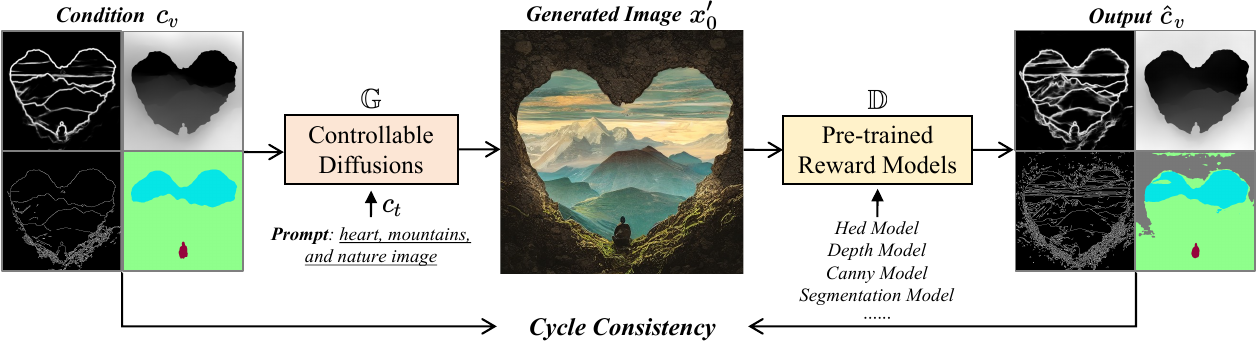}
    \vspace{-0.7cm}
    \caption{
        \textbf{Illustration of the cycle consistency}.  We first prompt the diffusion model $\mathbb{G}$ to generate an image \(x'_0\) based on the given image condition \(c_v\) and text prompt $c_t$, then extract the corresponding image condition $\hat{c}_v$ from the generated image $x'_0$ using pre-trained discriminative models $\mathbb{D}$. The cycle consistency is defined as the similarity between the extracted condition $\hat{c}_v$ and input condition $c_v$.
    }
    \vspace{-0.5cm}
    \label{fig:method_high-level}
\end{figure}

\vspace{-0.2cm}
\begin{itemize}[leftmargin=*]
\item \underline{\textit{New Insight}}: 
We reveal that existing efforts in controllable generation still perform poorly in terms of controllability, with generated images significantly deviating from input conditions and lacking a clear strategy for improvement.

\item \underline{\textit{Consistency Reward Feedback}}: 
We show that pre-trained discriminative models can serve as powerful visual reward models to improve the controllability of controllable diffusion models in a cycle-consistency manner.

\item \underline{\textit{Efficient Reward Fine-tuning}}: 
We disrupt the consistency between input images and conditions, and enable the single-step denoising for efficient reward fine-tuning, avoiding time and memory overheads caused by image sampling.

\item \underline{\textit{Evaluation and Promising Results}}:
We provide a unified and public evaluation of controllability under various conditional controls, and demonstrate that ControlNet++ comprehensively outperforms existing methods.
\end{itemize}

\section{Related Work}
\label{sec:related_work}
\vspace{-0.3cm}
\subsection{Diffusion-based Generative Models}
\vspace{-0.2cm}
The diffusion probabilistic model presented in~\cite{diffusion} has undergone substantial advancements~\cite{var_diffusion,diffusion_beat_gan,classifier-free}, thanks to iterative refinements in training and sampling strategies~\cite{ddpm,ddim,score-based_diffusion}. To alleviate the computational demands for training diffusion models, Latent Diffusion~\cite{sd} maps the pixel space diffusion process into the latent feature space. In the realm of text-to-image synthesis, diffusion models~\cite{sd,glide,imagen,sdxl,dalle,dalle2} integrate cross-attention mechanisms between UNet~\cite{unet} denoisers and text embeddings from pre-trained language models like CLIP~\cite{clip} and T5~\cite{t5} to facilitate reasonable text-to-image generation. Furthermore, diffusion models are employed across image editing tasks~\cite{textual_inversion,sdedit,instruct_p2p,imagic} by manipulating inputs~\cite{dalle2}, editing cross-attentions~\cite{p2p}, and fine-tuning models~\cite{db}. Despite the astonishing capabilities of diffusion models, language is a sparse and highly semantic representation, unsuitable for describing dense, low-semantic images. Furthermore, existing methods~\cite{sd,sdxl} still struggle to understand detailed text prompts, posing a severe challenge to the controllable generation~\cite{controlnet}. 

\begin{figure}[t!]\centering
    \includegraphics[width=1.0\linewidth]{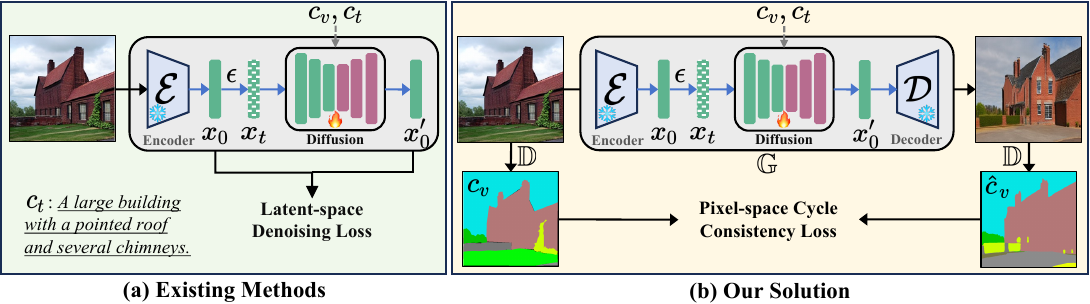}
    \vspace{-0.7cm}
    \caption{
    \textbf{(a)} Existing methods achieve implicit controllability by introducing image-based conditional control $c_v$ into the denoising process of diffusion models, with the guidance of latent-space denoising loss. \textbf{(b)} We utilize discriminative reward models $\mathbb{D}$ to explicitly optimize the controllability of $\mathbb{G}$ via pixel-level cycle consistency loss.
    }
    \vspace{-0.5cm}
    \label{fig:comparasion}
\end{figure}

\vspace{-0.2cm}
\subsection{Controllable Text-to-Image Diffusion Models}
\vspace{-0.2cm}
To achieve conditional control in pre-trained text-to-image diffusion models, ControlNet~\cite{controlnet} and T2I-Adapter~\cite{t2i_adapter} introduce additional trainable modules for guided image generation.
Furthermore, recent research employs various prompt engineering~\cite{gligen,control_by_gpt,reco} and cross-attention constraints~\cite{boxdiff,training-free-control,humansd} for a more regulated generation. Some methods also explore multi-condition or multi-modal generation within a single diffusion model~\cite{unicontrol,cocktail,unicontrolnet} or focus on the instance-based controllable generation~\cite{instance_diffusion,migc}. However, despite these methods exploring feasibility and applications, there still lacks a clear approach to enhance controllability under various controls. Furthermore, existing works implicitly learn controllability by the denoising process of diffusion models, while our ControlNet++ achieves this in an explicit cycle-consistency manner, as shown in Fig.~\ref{fig:comparasion}.

\vspace{-0.2cm}
\subsection{Linguistic and Visual Reward Models}
\vspace{-0.2cm}
The reward model is trained to evaluate how well the results of generative models align with human expectations, and its quantified results will be used to facilitate generative models for better and more controllable generation. It is usually trained with reinforcement learning from human feedback (RLHF) in NLP tasks~\cite{instruct_gpt,palm,llama2}, and has recently extended into the vision domain to improve the image quality for text-to-image diffusion models~\cite{alignsd,ddpo,dpok,imagereward,DRaFT,AlignProp}. However, image quality is an exceedingly subjective metric, fraught with individual preferences, and requires the creation of new datasets with human preferences~\cite{imagereward,alignsd,hpsv2,pickscore} and the training of reward models~\cite{imagereward,hpsv2,AlignProp}. Diverging from the pursuit of global image quality with subjective human preference in current research, we target the more fine-grained and objective goal of controllability.
Also, it's more cost-effective to obtain AI feedback compared to human feedback.


\section{Method}
In this section, we first introduce the background of diffusion models in Sec.~\ref{subsec:preliminary}. In Sec.~\ref{subsec:reward_control}, we discuss how to design the cycle consistency loss for controllable diffusion models to enhance the controllability. 
Finally, in Sec.~\ref{subsec:efficient_reward}, we examine the efficiency issues with the straightforward solution and correspondingly propose an efficient reward strategy that utilizes the single-step denoised images for consistency loss, instead of sampling images from random noise.

\subsection{Preliminary}
\label{subsec:preliminary}
The diffusion models~\cite{ddpm} define a Markovian chain of diffusion forward process $q(x_t | x_0)$ by gradually adding noise to input data $x_0$:
\begin{equation}
x_t=\sqrt{\bar{\alpha}_t} x_0+\sqrt{1-\bar{\alpha}_t} \epsilon, \quad \epsilon \sim \mathcal{N}(\mathbf{0}, {I}),
\label{eq:add_noise}
\end{equation}
where $\epsilon$ is a noise map sampled from a Gaussian distribution, and $\bar{\alpha}_t:=\prod_{s=0}^t \alpha_s$. $\alpha_t = 1 - \beta_t$ is a differentiable function of timestep $t$, which is determined by the denoising sampler such as DDPM~\cite{ddpm}. To this end, the diffusion training loss can be represented by:
\begin{equation}
\mathcal{L}\left(\epsilon_\theta\right)=\sum_{t=1}^T \mathbb{E}_{x_0 \sim q\left(x_0\right), \epsilon \sim \mathcal{N}(\mathbf{0}, {I})}\left[\left\|\epsilon_\theta\left(\sqrt{\bar{\alpha}_t} x_0+\sqrt{1-\bar{\alpha}_t} \epsilon\right)-\epsilon\right\|_2^2\right].
\end{equation}
In the context of controllable generation~\cite{controlnet,t2i_adapter}, with given image condition $c_v$ and text prompt $c_t$, the diffusion training loss at timestep $t$ can be re-written as:
\begin{equation}
\mathcal{L}_{\text{train}} = \mathbb{E}_{x_0, t, c_t, c_v, \epsilon \sim \mathcal{N}(0,1)} \left[ \| \epsilon_\theta\left(x_t, t, c_t, c_v\right) - \epsilon \| _2^2 \right].
\label{loss:diffusion}
\end{equation}

During the inference, given a random noise $x_T \sim \mathcal{N}(\mathbf{0}, \mathbf{I})$, we can predict final denoised image $x_0$ with the step-by-step denoising process~\cite{ddpm}:
\begin{equation}
x_{t-1}=\frac{1}{\sqrt{\alpha_t}}\left(x_t-\frac{1-\alpha_t}{\sqrt{1-\bar{\alpha}_t}} {\epsilon}_\theta\left(\mathbf{x}_t, t\right)\right)+\sigma_t \epsilon,
\end{equation}
where $\epsilon_\theta$ refers to the predicted noise at timestep $t$ by U-Net~\cite{unet} with parameters $\theta$, and $\sigma_t = \frac{1-\bar{\alpha}_{t-1}}{1-\bar{\alpha}_t} \beta_t$ is the variance of posterior Gaussian distribution $p_{\theta}(x_0)$.

\subsection{Reward Controllability with Consistency Feedback}
\label{subsec:reward_control}
As we model controllability as the consistency between input conditions and the generated images, we can naturally quantify this outcome through the discriminative reward models. Once we quantify the results of the generative model, we can perform further optimization for more controllable generation based on these quantified results in a unified manner for various conditional controls.

\begin{figure*}[t!]\centering
    \includegraphics[width=1.0\linewidth]{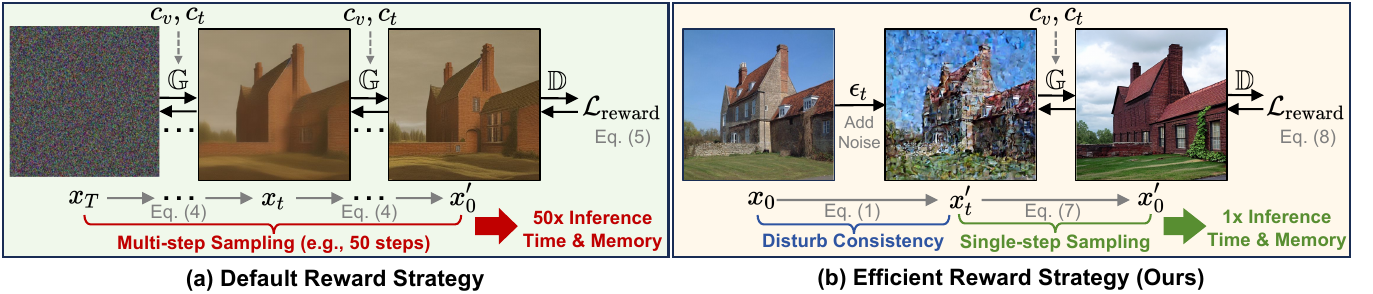}
    \vspace{-0.7cm}
    \caption{
    \textbf{(a)} Pipeline of default reward fine-tuning strategy. Reward fine-tuning requires sampling all the way to the full image. Such a method needs to keep all gradients for each timestep and the memory required is unbearable by current GPUs.
    \textbf{(b)} Pipeline of our efficient reward strategy. We add a small noise (\( t \) $\leq$ \( t_{thre} \)) to disturb the consistency between input images and conditions, then the single-step denoised image can be directly used for efficient reward fine-tuning. 
    }
    \vspace{-0.5cm}
    \label{fig:efficient_reward}
\end{figure*}

To be more specific, we minimize the consistency loss between the input condition $c_{v}$ and the corresponding output condition $\hat{c}_{v}$ of the generated image $x_0'$, as depicted in Fig.~\ref{fig:method_high-level}.
The reward consistency loss can be formulated as:

\begin{equation}
\begin{aligned}
\mathcal{L}_{\text {reward }} & =\mathcal{L}\left(c_v, \hat{c}_v\right) \\
& =\mathcal{L}\left(c_v, \mathbb{D}\left(x_0'\right)\right) \\
& =\mathcal{L}\left(c_v, \mathbb{D}\left[\mathbb{G}^{T}\left(c_t, c_v, x_T, t\right)\right]\right),
\label{eq:default_reward}
\end{aligned}
\end{equation}
where $\mathbb{G}^T\left(c_t, c_v, x_T, t\right)$ denotes the process that the model performs \textbf{ $T$ denoising steps } to generate the image $x_0'$ from random noise $x_T$, as shown in the Fig.~\ref{fig:efficient_reward} (a). 
Here, $\mathcal{L}$ is an abstract metric function that can take on different concrete forms for different visual conditions. For example, in the context of using segmentation mask as the input conditional control, $\mathcal{L}$ could be the per-pixel cross-entropy loss. The reward model $\mathcal{D}$ is also dependent on the condition, and we use the UperNet~\cite{upernet} for segmentation mask conditions.
The details of loss functions and reward models are summarized in the supplementary material. 

In addition to the reward loss, we also employ diffusion training loss in Eq.~\ref{loss:diffusion} to ensure that the original image generation capability is not compromised since they have different optimization goals. Finally, the total loss is the combination of $\mathcal{L}_{\text{train}}$ and $\mathcal{L}_{\text{reward}}$:
\begin{equation}
\mathcal{L}_{\text{total}} = \mathcal{L}_{\text{train}} + \lambda \cdot \mathcal{L}_{\text{reward}},
\end{equation}
where $\lambda$ is a hyper-parameter to adjust the weight of the reward loss. Through this approach, the consistency loss can guide the diffusion model on how to sample at different timesteps to obtain images more consistent with the input controls, thereby enhancing controllability. 
Nonetheless, directly applying such reward consistency still poses challenges in efficiency in real-world settings.

\subsection{Efficient Reward Fine-tuning}
\label{subsec:efficient_reward}

To achieve the pixel-space consistency loss $\mathcal{L}_{\text{reward}}$, it requires $x_0$, the final diffused image, to calculate the reward consistency from the reward models. As modern diffusion models, such as Stable Diffusion~\cite{sd}, require multiple steps, \eg, 50 steps, to render a full image, directly using such a solution is impractical in realistic settings:
(1) multiple time-consuming samplings are required to derive images from random noise. (2) to enable gradient backpropagation, we have to store gradients at each timestep, meaning the GPU memory usage will increase linearly with the number of time-steps. Taking ControlNet as an example, when the batch size is 1 with FP16 mixed precision, the GPU memory required for a single denoising step and storing all training gradients is approximately 6.8GB. If we use the 50-step inference with the DDIM~\cite{ddim} scheduler, approximately 340GB of memory is needed to perform reward fine-tuning on a single sample, which is nearly impossible to achieve with current hardware capabilities. Although GPU memory consumption can be reduced by employing techniques such as Low-Rank Adaptation (LoRA)~\cite{lora,DRaFT}, gradient checkpointing~\cite{gradient_checkpoint,DRaFT}, or stop-gradient~\cite{imagereward}, the efficiency degradation caused by the number of sampling steps required to generate images remains significant and cannot be overlooked. Therefore, an efficient reward fine-tuning approach is necessary.


In contrast to diffusing from random noise $x_T$ to obtain the final image $x_0$, as illustrated in Fig.~\ref{fig:efficient_reward} (a), we instead propose an  \textit{one-step} efficient reward strategy.
Specifically, instead of randomly sampling from noise, we add noise to the training images $x_0$, thereby explicitly disturbing the consistency between the diffusion inputs $x'_t$ and their conditional controls $c_v$, by performing diffusion forward process $q(x_t|x_0)$ in Eq.~\ref{eq:add_noise}. We demonstrate this process as the \textcolor[RGB]{53,84,159}{\textbf{Disturb Consistency}} in Fig.~\ref{fig:efficient_reward} (b)
, which is the same procedure as the standard diffusion training process. 
When the added noise $\epsilon$ is relatively small, we can predict the original image $x'_0$ by performing single-step sampling\footnote{We provide a more detailed proof in the supplementary material.} on disturbed image $x'_t$~\cite{ddpm}:
\begin{equation}
x_0 \approx x'_0=\frac{x'_{t}-\sqrt{1-\alpha_{t}} \epsilon_\theta\left(x'_{t}, c_v, c_t, t\right)}{\sqrt{\alpha_{t}}},
\end{equation}
and then we directly utilize the denoised image $x'_0$ to perform reward fine-tuning:
\begin{equation}
\begin{aligned}
\mathcal{L}_{\text{reward}} &= \mathcal{L}({c}_{v}, \hat{c}_{v})=\mathcal{L}({c}_{v}, \mathbb{D}(x'_0))=\mathcal{L}({c}_{v}, \mathbb{D}[\mathbb{G}(c_{t}, {c}_{v}, x'_t, t)]).
\end{aligned}
\label{eq:reward_loss}
\end{equation}
Essentially, the process of adding noise destroys the consistency between the input image and its condition. Then the reward fine-tuning in Eq. \ref{eq:reward_loss} instructs the diffusion model to generate images that can reconstruct the consistency, thus enhancing its ability to follow conditions during generation.

Please note that here we avoid the sampling process in Eq.~\ref{eq:default_reward}. Finally, the loss is the combination of diffusion training loss and the reward loss:
\begin{equation}
\mathcal{L}_{\text {total }}= \begin{cases}\mathcal{L}_{\text {train }}+\lambda \cdot \mathcal{L}_{\text {reward }}, & \text {if } t \leq t_{\text {thre},} \\ \mathcal{L}_{\text {train }}, & \text {otherwise, }\end{cases}
\end{equation}
where $t_{thre}$ denotes the timestep threshold, which is a hyper-parameter used to determine whether a noised image $x_t$ should be utilized for reward fine-tuning. We note that a small noise $\epsilon$ (\ie, a relatively small timestep $t$) can disturb the consistency and lead to effective reward fine-tuning. When the timestep $ t $ is large, $ x_t $ is closer to the random noise $ x_T $, and predicting $ x'_0 $ directly from $ x_t $ results in severe image distortion. The advantage of our efficient rewarding is that $x_t$ can be employed both to train and reward the diffusion model without the need for time and GPU memory costs caused by multiple sampling, thereby significantly improving the efficiency during the reward fine-tuning stage.

During the reward fine-tuning phases, we freeze the pre-trained discriminative reward model and text-to-image model, and only update the ControlNet following its original implementation, which ensures the generative capabilities are not compromised. We also observe that using only the reward loss will lead to image distortion, aligning with the conclusions drawn in previous studies~\cite{imagereward}.

\section{Experiments}

\subsection{Experimental Setup}
\label{subsec:setup}
\paragraph{Condition Controls and Datasets.}
Given that existing text-image paired datasets for generative models are unable to provide accurate conditional control data pairs~\cite{laion400m,laion5b}, such as image-segmentation pairs, we endeavor to select specific datasets for different tasks that can offer more precise image-label data pairs. More specifically, ADE20K~\cite{ade20k_0,ade20k_1} and COCOStuff~\cite{cocostuff} are used for the segmentation mask condition following ControlNet~\cite{controlnet}. For the canny edge map, hed edge map, lineart map, and depth map condition, we utilize the MultiGen-20M dataset proposed by UniControl~\cite{unicontrol}, which is a subset of LAION-Aesthetics~\cite{laion5b}. For the datasets without text caption such as ADE20K, we utilize MiniGPT-4~\cite{minigpt4} to generate the image caption with the instruction ``\textit{Please briefly describe this image in one sentence}''. The training and inference resolution is 512$\times$512 for all datasets and methods. Details are provided in the supplementary material.

\vspace{-0.2cm}
\paragraph{Evaluation and Metrics.}
We train ControlNet++ on the training set of each corresponding dataset and evaluate all methods on the validation dataset. All the experiments are evaluated under 512$\times$512 resolution for fair comparison. For each condition, we evaluate the controllability by measuring the similarity between the input conditions and the extracted conditions from generated images of diffusion models. For semantic segmentation and depth map controls, we use mIoU and RMSE as evaluation metrics respectively, which is a common practice in related research fields. For the edge task, we use F1-Score for hard edges (canny edge) because it can be regarded as a binary classification problem of 0 (non-edge) and 1 (edge) and has a serious long-tail distribution, following the standard evaluation in edge detection~\cite{edge_det}. The threshold used for evaluation is (100, 200) for OpenCV, and (0.1, 0.2) for Kornia implementation. The SSIM metric is used for the soft edges conditional controls (\ie, hed edge \& lineart edge) following previous works~\cite{unicontrolnet}. For ControlNet++, we use the UniPC~\cite{unipc} sampler with 20 denoising steps to generate images with the original text prompt following ControlNet v1.1~\cite{controlnet}, without any negative prompts. 
For other methods beyond ControlNet and ours, we utilized their open-source code to generate images and conducted fair evaluations under the same data, without changing their inference configures such as the number of inference steps or denoising sampler.

\vspace{-0.2cm}
\paragraph{Baselines.}
Our evaluation primarily focuses on T2I-Adapter~\cite{t2i_adapter}, ControlNet v1.1~\cite{controlnet}, GLIGEN~\cite{gligen}, Uni-ControlNet~\cite{unicontrolnet}, and UniControl~\cite{unicontrol}, as these methods are pioneering in the realm of controllable text-to-image diffusion models and offer public model weights for various image conditions. To ensure fairness of evaluation, all methods use the same image conditions and text prompts. While most methods employ the user-friendly SD1.5 as their text-to-image model for controllable generation, we have observed that recently there are a few models based on SDXL~\cite{sdxl}. Therefore, we also report the controllability results for ControlNet-SDXL and T2I-Adapter-SDXL. Please note that ControlNet-SDXL mentioned here is not an officially released model as in ControlNet~\cite{controlnet}.

\begin{table}[t!]
\caption{Controllability comparison with state-of-the-art methods under different conditional controls and datasets. $\uparrow$ denotes higher result is better, while $\downarrow$ means lower is better. ControlNet++ achieves significant controllability improvements. `-' indicates that the method does not provide a public model for testing. We generate four groups of images in png format and report the average result to reduce random errors.}
\vspace{-0.4cm}
\resizebox{\textwidth}{!}{%
\begin{tabular}{c|c|c|c|c|c|c|c}
\toprule[2pt]
\multicolumn{1}{c|}{\textbf{\begin{tabular}[c]{@{}c@{}}Condition\\ (Metric)\end{tabular}}} & \multicolumn{1}{c|}{} & \multicolumn{2}{c|}{\textbf{\begin{tabular}[c]{@{}c@{}}Seg. Mask\\ (mIoU $\uparrow$)\end{tabular}}} & \multicolumn{1}{c|}{\textbf{\begin{tabular}[c]{@{}c@{}}Canny Edge\\ (F1 Score $\uparrow$)\end{tabular}}} & \multicolumn{1}{c|}{\textbf{\begin{tabular}[c]{@{}c@{}}Hed Edge\\ (SSIM $\uparrow$)\end{tabular}}} & \multicolumn{1}{c|}{\textbf{\begin{tabular}[c]{@{}c@{}}LineArt Edge\\ (SSIM $\uparrow$)\end{tabular}}} & \multicolumn{1}{c}{\textbf{\begin{tabular}[c]{@{}c@{}}Depth Map\\ (RMSE $\downarrow$)\end{tabular}}} \\ \cline{1-1} \cline{3-8} 
\multicolumn{1}{c|}{\textbf{Dataset}} & \multicolumn{1}{c|}{\multirow{-3}{*}{\textbf{\begin{tabular}[c]{@{}c@{}}T2I\\Model\end{tabular}}}} & \multicolumn{1}{c|}{\textbf{ADE20K}} & \multicolumn{1}{c|}{\textbf{COCO-Stuff}} & \multicolumn{1}{c|}{\textbf{MultiGen-20M}} & \multicolumn{1}{c|}{\textbf{MultiGen-20M}} & \multicolumn{1}{c|}{\textbf{MultiGen-20M}} & \multicolumn{1}{c}{\textbf{MultiGen-20M}} \\
\hline
ControlNet & SDXL & - & - & - & - & - & 40.00 \\
T2I-Adapter & SDXL & - & - & 28.01 & - & 0.6394 & 39.75 \\
\hline
T2I-Adapter & SD1.5 & 12.61 & - & 23.65 & - & - & 48.40 \\
Gligen & SD1.4 & 23.78 & - & 26.94 & 0.5634 & - & 38.83 \\
Uni-ControlNet & SD1.5 & 19.39 & - & 27.32 & 0.6910 & - & 40.65 \\
UniControl & SD1.5 & 25.44 & - & 30.82 & 0.7969 & - & 39.18 \\
ControlNet & SD1.5 & 32.55 & 27.46 & 34.65 & 0.7621 & 0.7054 & 35.90 \\ 
\textbf{Ours} & SD1.5 & {\textbf{43.64}} & {\textbf{34.56}} & {\textbf{37.04}} & {\textbf{0.8097}} & {\textbf{0.8399}} & {\textbf{28.32}}
\\ \bottomrule[2pt]
\end{tabular}%
}
\vspace{-0.2cm}

\label{tab:controllability}
\end{table}

\subsection{Experimental Results}
\label{subsec:our_results}
\paragraph{Comparison of Controllability with State-of-the-art Methods.}
The experimental results are shown in Tab.~\ref{tab:controllability}, which can be summarized as the following observations:
(1) Existing methods still underperform in terms of controllability, struggling to achieve precise controlled generation. For instance, current methods (\ie, ControlNet) achieve only a 32.55 mIoU for images generated under the condition of segmentation masks, which is far from its performance on real datasets with a 50.7 mIoU, under the same evaluation from Mask2Former segmentation model~\cite{mask2former}.
(2) Our ControlNet++ significantly outperforms existing works in terms of controllability across various conditional controls. For example, it achieves 11.1\% RMSE improvements against previous state-of-the-art methods for the depth map condition;
(3) For controllable diffusion models, the strength of the text-to-image backbone does not affect its controllability. As shown in the table, although SDXL-based~\cite{sdxl} ControlNet and T2I-Adapter have better controllability on some specific tasks, the improvement is not large and is not significantly better than the counterparts with SD 1.5~\cite{sd}.

\begin{table}[t!]
\caption{FID ($\downarrow$) comparison with state-of-the-art methods under different conditional controls and datasets. All the results are conducted on 512$\times$512 image resolution with Clean-FID implementation~\cite{clean-fid} for fair comparisons. `-' indicates that the method does not provide a public model for testing. We generate four groups of images in png format and report the average result to reduce random errors.}
\vspace{-0.4cm}
\resizebox{\textwidth}{!}{%
\begin{tabular}{c|c|c|c|c|c|c|c}
\toprule[2pt]
\multicolumn{1}{c|}{} & \multicolumn{1}{c|}{} & \multicolumn{2}{c|}{\textbf{Seg. Mask}} & \multicolumn{1}{c|}{\textbf{Canny Edge}} & \multicolumn{1}{c|}{\textbf{Hed Edge}} & \multicolumn{1}{c|}{\textbf{LineArt Edge}} & \textbf{Depth Map} \\ \cline{3-8} 
\multicolumn{1}{c|}{\multirow{-2}{*}{\textbf{Method}}} & {\multirow{-2}{*}{\textbf{\begin{tabular}[c]{@{}c@{}}T2I\\Model\end{tabular}}}} & \multicolumn{1}{c|}{\textbf{ADE20K}} & \multicolumn{1}{c|}{\textbf{COCO}} & \multicolumn{1}{c|}{\textbf{MultiGen-20M}} & \multicolumn{1}{c|}{\textbf{MultiGen-20M}} & \multicolumn{1}{c|}{\textbf{MultiGen-20M}} & \textbf{MultiGen-20M} \\ \hline
Gligen & SD1.4 & 33.02 & - & 18.89 & - & - & 18.36 \\
T2I-Adapter & SD1.5 & 39.15 & - & 15.96 & - & - & 22.52 \\
UniControlNet & SD1.5 & 39.70 & - & 17.14 & 17.08 & - & 20.27 \\
UniControl & SD1.5 & 46.34 & - & 19.94 & 15.99 & - & 18.66 \\
ControlNet & SD1.5 & 33.28 & 21.33 & \textbf{14.73} & 15.41 & 17.44 & 17.76 \\
\textbf{Ours} & SD1.5 & \textbf{29.49} & \textbf{19.29} & 18.23 & \textbf{15.01} & \textbf{13.88} & \textbf{16.66}
\\ \bottomrule[2pt]
\end{tabular}%
}
\vspace{-0.4cm}

\label{tab:fid}
\end{table}

\begin{table}[t!]
\caption{CLIP-score ($\uparrow$) comparison with state-of-the-art methods under different conditional controls and datasets. `-' indicates that the method does not provide a public model for testing. We generate four groups of images in png format and report the average result to reduce random errors.}
\vspace{-0.4cm}
\resizebox{\textwidth}{!}{%
\begin{tabular}{c|c|c|c|c|c|c|c}
\toprule[2pt]
\multicolumn{1}{c|}{} & \multicolumn{1}{c|}{} & \multicolumn{2}{c|}{\textbf{Seg. Mask}} & \multicolumn{1}{c|}{\textbf{Canny Edge}} & \multicolumn{1}{c|}{\textbf{Hed Edge}} & \multicolumn{1}{c|}{\textbf{LineArt Edge}} & \textbf{Depth Map} \\ \cline{3-8} 
\multicolumn{1}{c|}{\multirow{-2}{*}{\textbf{Method}}} & {\multirow{-2}{*}{\textbf{\begin{tabular}[c]{@{}c@{}}T2I\\Model\end{tabular}}}} & \multicolumn{1}{c|}{\textbf{ADE20K}} & \multicolumn{1}{c|}{\textbf{COCO}} & \multicolumn{1}{c|}{\textbf{MultiGen-20M}} & \multicolumn{1}{c|}{\textbf{MultiGen-20M}} & \multicolumn{1}{c|}{\textbf{MultiGen-20M}} & \textbf{MultiGen-20M} \\ \hline
Gligen & SD1.4 & 31.12 & - & 31.77 & - & - & 31.75 \\
T2I-Adapter & SD1.5 & 30.65 & - & 31.71 & - & - & 31.46 \\
UniControlNet & SD1.5 & 30.59 & - & 31.84 & 31.94 & - & 31.66 \\
UniControl & SD1.5 & 30.92 & - & 31.97 & 32.02 & - & \textbf{32.45} \\
ControlNet & SD1.5 & 31.53 & \textbf{13.31} & \textbf{32.15} & \textbf{32.33} & \textbf{32.46} & \textbf{32.45} \\
\textbf{Ours} & SD1.5 & \textbf{31.96} & 13.13 & 31.87 & 32.05 & 31.95 & 32.09
\\ \bottomrule[2pt]
\end{tabular}%
}
\vspace{-0.4cm}
\label{tab:clip_score}
\end{table}

\vspace{-0.2cm}
\paragraph{Comparison of Image Quality with State-of-the-art Methods.}
To verify whether improving controllability leads to a decline in image quality, we reported the FID (Fréchet Inception Distance) metrics of different methods under various conditional generation tasks in Tab.~\ref{tab:fid}. We discovered that, compared to existing methods, ControlNet++ generally exhibits superior FID values in most cases, indicating that our approach, while enhancing the controllability of conditional controls, does not result in a decrease in image quality. This can also be observed in Fig.~\ref{fig:visualization_comparison_1}. We provide more visual examples in the supplementary material. 

\vspace{-0.2cm}
\paragraph{Comparison of CLIP score with State-of-the-art Methods.}
Our ControlNet++ aims to improve the controllability of diffusion models using image-based conditions. Concerned about the potential adverse effects on text controllability, we evaluated various methods using CLIP-Score metrics across different datasets to measure the similarity between generated images and input text. As indicated in Tab.~\ref{tab:clip_score}, ControlNet++ achieved comparable or superior CLIP-Score outcomes on several datasets relative to existing approaches. This suggests that our method not only markedly enhances conditional controllability but also preserves the original model's text-to-image generation proficiency.





\begin{figure*}[t!]\centering
    \includegraphics[width=0.8\columnwidth]{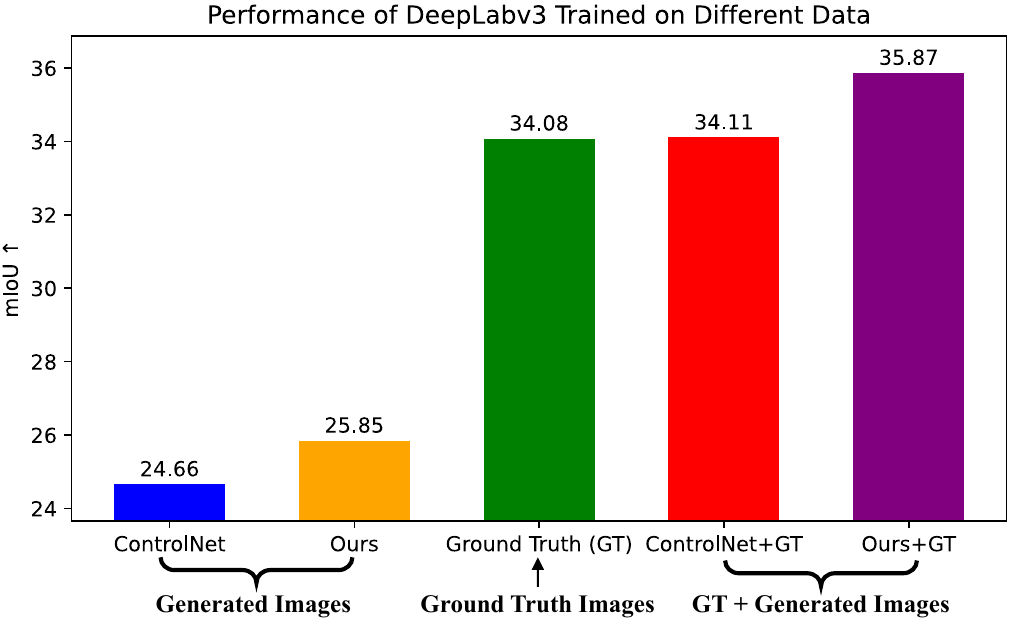}
    \vspace{-0.2cm}
    \caption{
        Training DeepLabv3 (MobileNetv2) from scratch with different images, including ground truth images from ADE20K, and the generated images from ControlNet and ours. All the labels (\ie, segmentation masks) are ground truth labels in ADE20K. \textbf{Please note improvements here are non-trivial for semantic segmentation.}
    }
    \vspace{-0.6cm}
    \label{fig:seg_training}
\end{figure*}

\paragraph{Effectiveness of Generated Images.}
To further validate our improvements in controllability and their impact, we use the generated images along with real human-annotated labels to create a new dataset for training discriminative models from scratch. Please note that the only difference from the original dataset used to train the discriminative model is that we have replaced the images with those generated by the controllable diffusion model while keeping the labels unchanged. If the generative model exhibits good controllability, the quality of the constructed dataset will be higher, thereby enabling to train a stronger model.

Specifically, we conduct experiments on the ADE20K~\cite{ade20k_0,ade20k_1} dataset on DeepLabv3 with MobileNetv2 backbone~\cite{deeplabv3}. We use the standard training dataset (20210 training samples) to train the discriminative model and the validation dataset (5000 evaluation samples) for evaluation. We show the experimental results in Fig.~\ref{fig:seg_training}, the segmentation model trained on our images outperforms the baseline results (ControlNet) by 1.19 mIoU. \textbf{Please note that this improvement is significant in segmentation tasks}. For instance, Mask2Former~\cite{mask2former} improves previous SOTA MaskFormer~\cite{maskformer} with around 1.1 mIoU in semantic segmentation. In addition to conducting experiments solely on the generated dataset, we also combined generated data with real data to train the segmentation model. The experimental results indicate that augmenting real ground truth data with data generated by ControlNet does not yield additional performance improvements (34.11 v.s. 34.08). In contrast, augmenting real data with our generated data results in significant performance enhancements (+1.76 mIoU).

\begin{figure*}[t!]\centering
    \includegraphics[width=0.95\linewidth]{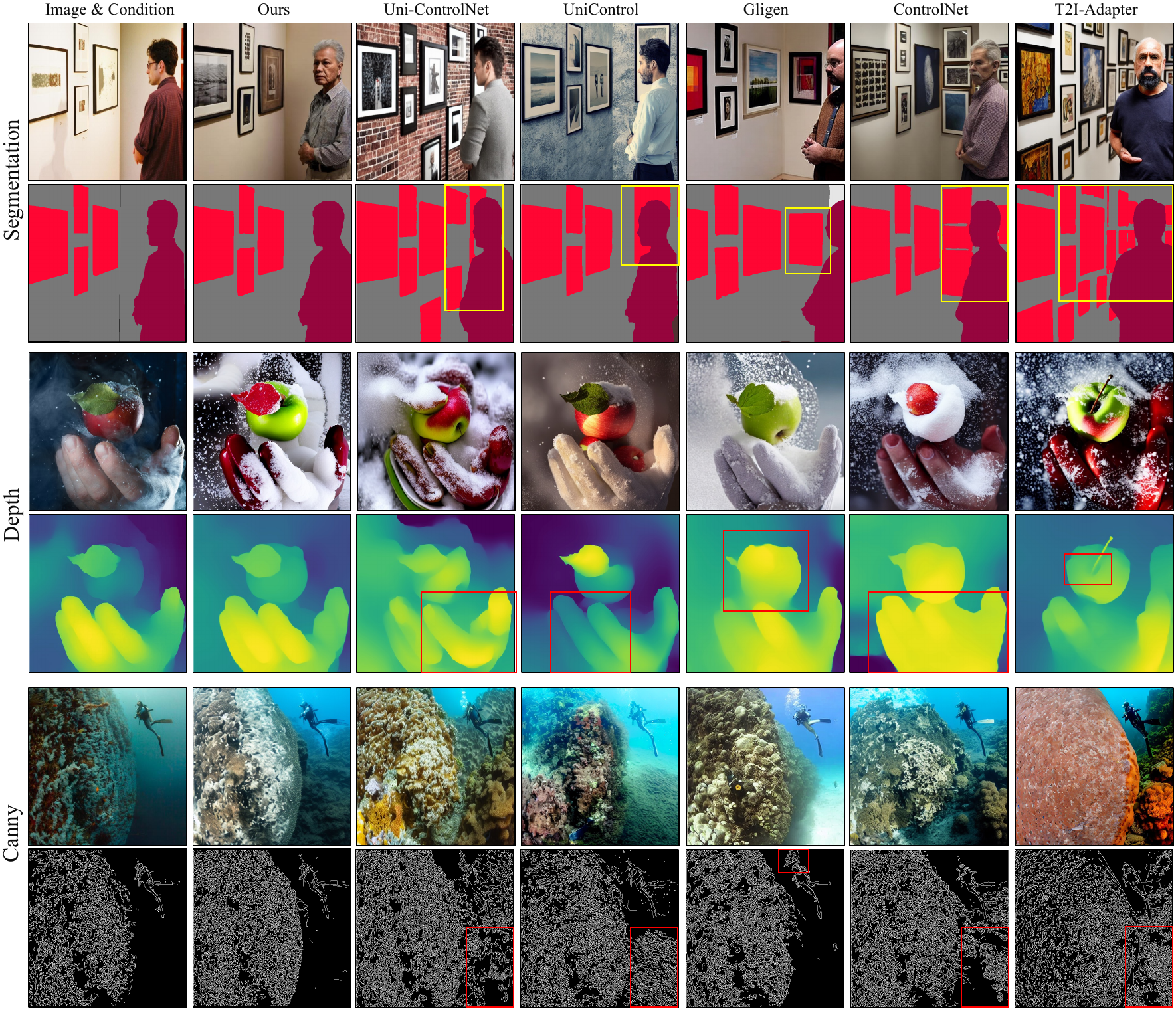}
    \vspace{-0.1cm}
    \includegraphics[width=0.95\linewidth]{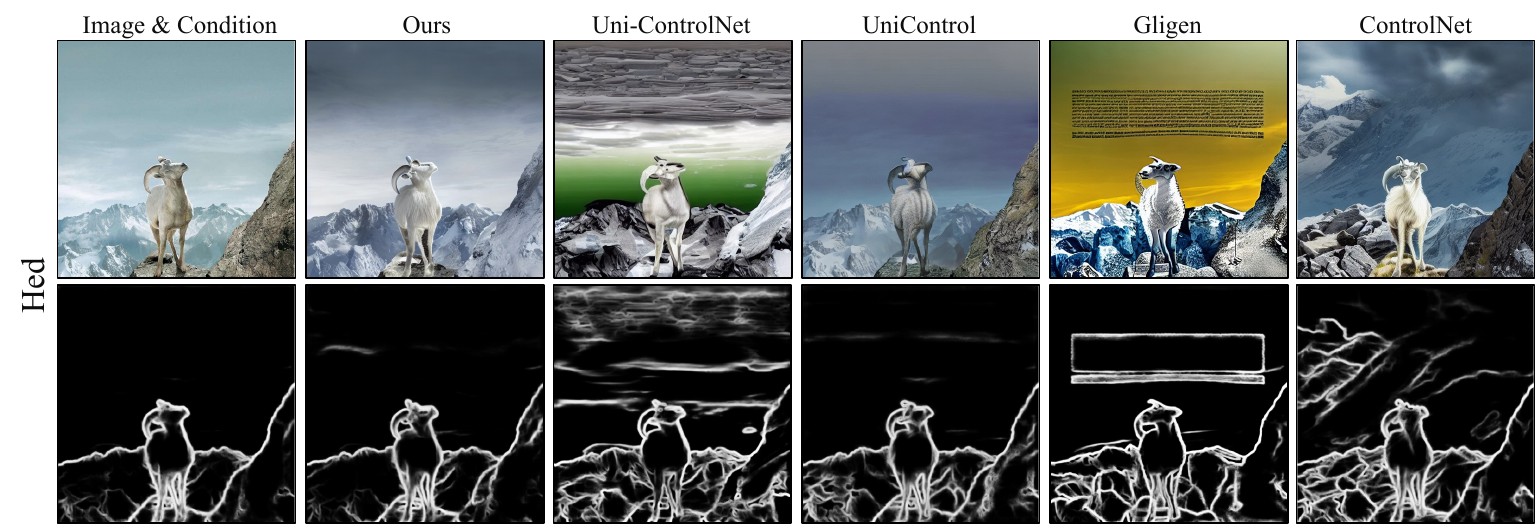}
    \vspace{-0.25cm}
    \caption{
    Visualization comparison results in different conditional controls.
    }
    \vspace{-0.5cm}
    \label{fig:visualization_comparison_1}
\end{figure*}

\begin{wrapfigure}{r}{0.45\textwidth}\centering
    \vspace{-0.5cm}
    \includegraphics[width=1.0\linewidth]{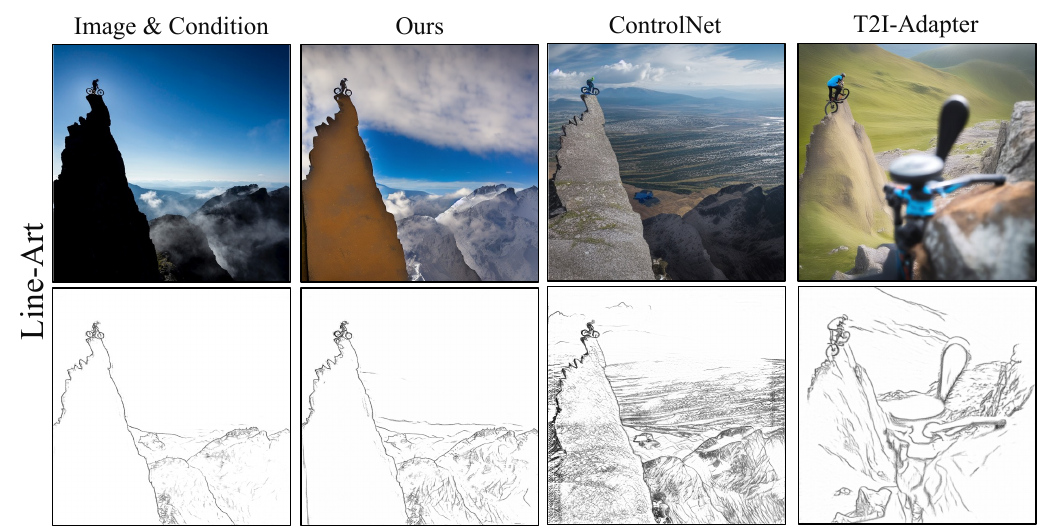}
    \vspace{-0.7cm}
    \caption{
    Comparison on Line-Art Edge. 
    }
    \vspace{-1.0cm}
    \label{fig:visualization_comparison_2}
\end{wrapfigure}

\paragraph{Qualitative Comparison.}
Figs.~\ref{fig:visualization_comparison_1} and \ref{fig:visualization_comparison_2} provide a qualitative comparison between our ControlNet++ and previous state-of-the-art methods across different conditional controls. When given the same input text prompts and image-based conditional controls, we observe that existing methods often generate areas
inconsistent with the image conditions. For instance, in the segmentation mask generation task, other methods often produce extraneous picture frames on the walls, resulting in a mismatch between the segmentation masks extracted from the generated image and the inputs. A similar situation occurs under depth conditions, where other methods fail to accurately represent the depth of different fingers. In contrast, images generated by ControlNet++ maintain good consistency with the input depth map.

\subsection{Ablation Study}
\label{subsec:ablation}

\begin{figure*}[h!]\centering
    \vspace{-0.7cm}
    \includegraphics[width=1.0\linewidth]{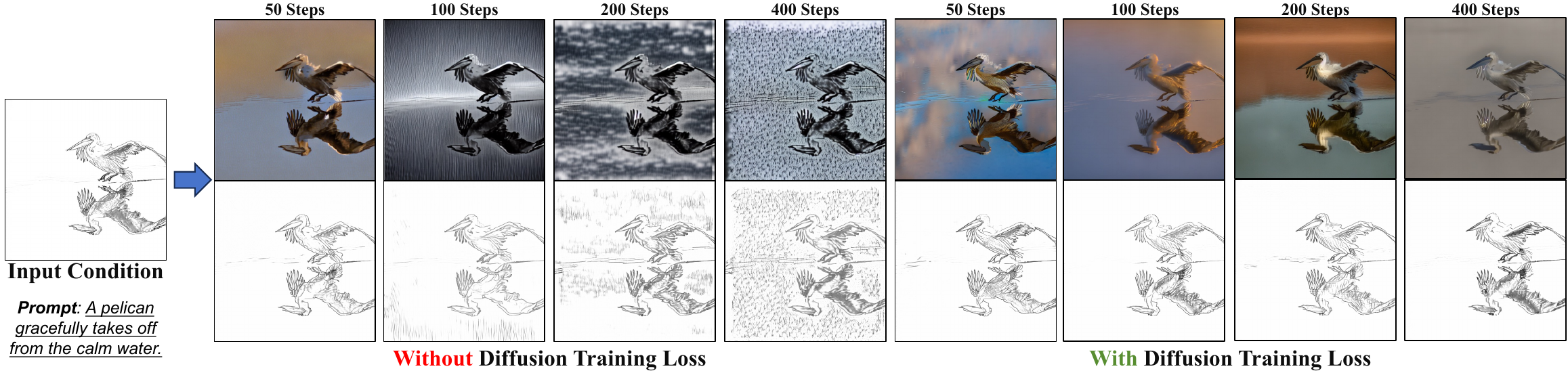}
    \vspace{-0.7cm}
    \caption{
    Ablation study on different loss settings during training. Using only pixel-level consistency loss leads to severe image distortion, affecting both image quality and controllability. However, when combined with diffusion training loss, it is possible to gradually improve controllability without compromising image quality.
    }
    \vspace{-0.8cm}
    \label{fig:ablation_loss}
\end{figure*}

\paragraph{Loss Settings.}
In Fig.~\ref{fig:ablation_loss}, we find that maintaining the original diffusion training process is crucial for preserving the quality and controllability of generated images. Relying solely on pixel-level consistency loss leads to severe image distortion, whereas training the model with both this loss and the diffusion training loss can enhance controllability without affecting image quality.

\paragraph{Generalizability of Efficient Reward Fine-tuning.}
Although the reward fine-tuning is used in a small subset of timesteps, it updates all the parameters of the ControlNet and therefore helps more timesteps to improve controllability during sampling. To prove this, we divide the sampling process into two parts: the unoptimized timesteps $[T, t_{thre}]$, and the optimized timesteps $[t_{thre}, 1]$ and use ControlNet and our model for inference crossly, with 20-step sampling following \href{https://huggingface.co/lllyasviel/control_v11p_sd15_canny/blob/115a470d547982438f70198e353a921996e2e819/control_net_canny.py#L34}{ControlNet}. Table~\ref{tab:ablation_timesteps} shows that our reward finetuning performed on a small number of timesteps $[t_{thre}, 1]$ can be generalized to larger timesteps $[T, t_{thre}]$.
\vspace{-0.7cm}
\begin{table}[h!]
\centering
\begin{minipage}[t]{0.45\linewidth}
    \centering
    \vspace{0.5cm}
    \caption{The impact of efficient reward fine-tuning on different timesteps.}
    \vspace{-0.4cm}
    \label{tab:ablation_timesteps}
    \resizebox{\linewidth}{!}{%
    \begin{tabular}{c|c|c}
    \toprule
    \textbf{\begin{tabular}[c]{@{}c@{}}Unoptimized\\ $[T, t_{thre}]$\end{tabular}} & \textbf{\begin{tabular}[c]{@{}c@{}}Optimized\\ $[t_{thre}, 1]$\end{tabular}} & \textbf{\begin{tabular}[c]{@{}c@{}}ADE20K\\ mIoU ($\uparrow$)\end{tabular}} \\ \hline
    ControlNet & ControlNet & 32.55 \\ \hline
    ControlNet & Ours & 38.03 \\ \hline
    Ours & ControlNet & 41.46 \\ \hline
    \textbf{Ours} & \textbf{Ours} & \textbf{43.64} \\
    \bottomrule
    \end{tabular}
    }
\end{minipage}
\hspace{0.01\linewidth}
\begin{minipage}[t]{0.52\linewidth}
    \centering
    \vspace{0.5cm}
    \caption{Stronger reward model (UperNet-R50) leads to better controllability than the weaker reward model (DeepLabv3-MBv2).}
    \vspace{-0.4cm}
    \resizebox{\linewidth}{!}{%
    \begin{tabular}{c|c|c}
        \toprule
        \textbf{Reward Model (RM)} & \textbf{RM mIoU$\uparrow$} & \textbf{Eval mIoU$\uparrow$} \\
        \midrule
        - & - & 32.55 \\
        DeepLabv3-MBv2 & 34.02 & 31.96 \\
        FCN-R101 & 39.91 & 40.44 \\
        UperNet-R50 & \textbf{42.05} & \textbf{43.64} \\
        \bottomrule
    \end{tabular}
    }
    \label{subtab:diff_reward_models}
\end{minipage}
\end{table}
\vspace{-0.45cm}

\paragraph{Choice of Different Reward Models.}
We demonstrate the effectiveness of different reward models in Tab.~\ref{subtab:diff_reward_models}, all the evaluation results (\ie, Eval mIoU in the table) are evaluated by the most powerful segmentation model Mask2Former~\cite{mask2former} with 56.01 mIoU, on ADE20K dataset. We experiment with three different reward models, including DeepLabv3~\cite{deeplabv3} with MobileNetv2~\cite{mbv2} backbone (DeepLabv3-MBv2), FCN~\cite{fcn} with ResNet-101~\cite{resnet} backbone (FCN-R101) and UperNet~\cite{upernet} with ResNet-50 backbone. The results demonstrate that a more powerful reward model leads to better controllability for controllable diffusion models.

\section{Discussion}
\vspace{-0.3cm}
\paragraph{\uline{How to make Hed/LineArt Edge 
 extraction methods differentiable?}}
The \href{https://github.com/lllyasviel/ControlNet-v1-1-nightly/blob/e2b44154b72965c5e11b1ccee941d550682e4701/annotator/hed/__init__.py#L56}{Hed} and \href{https://github.com/lllyasviel/ControlNet-v1-1-nightly/blob/e2b44154b72965c5e11b1ccee941d550682e4701/annotator/lineart/__init__.py#L94}{LineArt} Edge extraction models are neural networks without non-differentiable operations. Differentiability can be achieved by simply modifying forward code.

\vspace{-0.3cm}
\paragraph{\uline{Some conditions (e.g., Box/Sketch/Pose) are not available.}}
Our reward fine-tuning leverages a pre-trained ControlNet and a differentiable reward model. Currently, pre-trained ControlNet for object bounding boxes and differentiable reward models for sketches are lacking. In existing pose models, there are non-differentiable operations such as the NMS and keypoints grouping. We leave the question of how to extend consistency reward to more conditions to future work.

\begin{figure*}[h!]\centering
\vspace{-1.0cm}
    \includegraphics[width=1.0\linewidth]{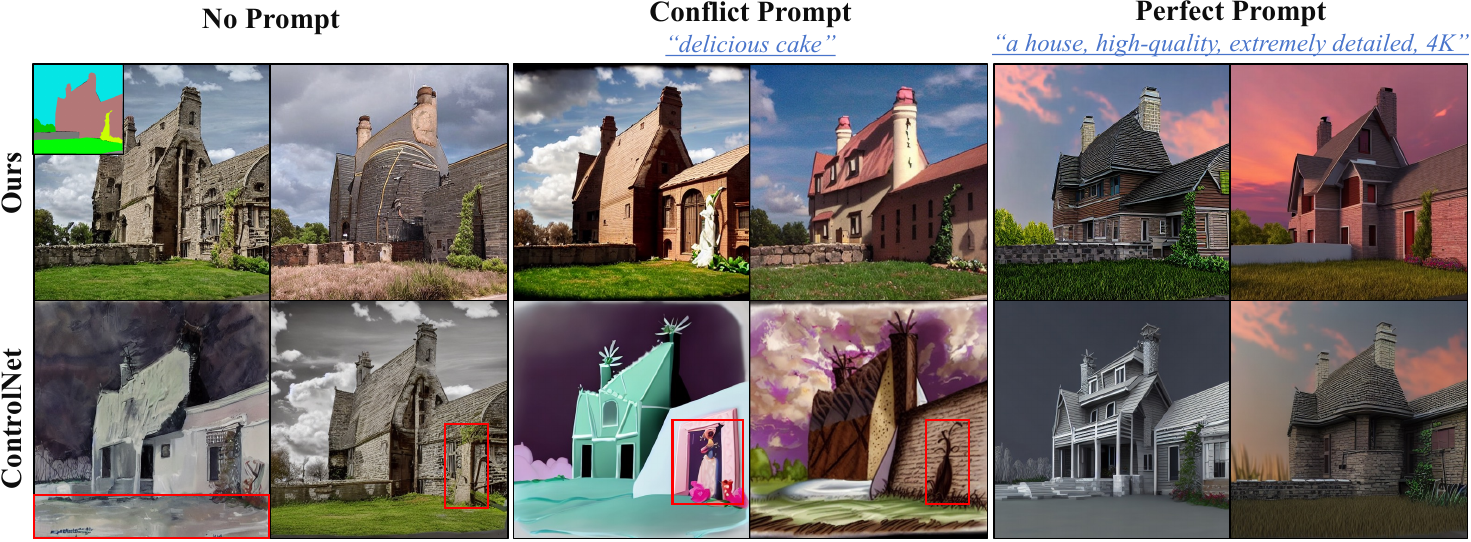}
    \vspace{-0.7cm}
    \caption{
    When the input text prompt is empty or conflicts with the image-based conditional controls (the segmentation map in the top left corner), ControlNet struggles to generate correct content (red boxes), whereas our method manages to generate it well.
    }
    \vspace{-1.0cm}
    \label{fig:ablation_prompt}
\end{figure*}

\vspace{-0.3cm}
\paragraph{\uline{Influence of Text Prompt.}}
We discuss how different types of text prompts (No Prompt, Conflicting Prompt, and Perfect Prompt) affect the final results. As shown in Fig.~\ref{fig:ablation_prompt}, when the text prompt is empty or there is a semantic conflict with the image conditional control, ControlNet often struggles to generate accurate content. In contrast, our ControlNet++ manages to generate images that comply with the input conditional controls under various text prompt scenarios.

\vspace{-0.1cm}
\section{Conclusion}
\vspace{-0.1cm}
In this paper, we demonstrate from both quantitative and qualitative perspectives that existing works focusing on controllable generation still fail to achieve precise conditional control, leading to inconsistency between generated images and input conditions. To address this issue, we introduce ControlNet++, it explicitly optimizes the consistency between input conditions and generated images using a pre-trained discriminative reward model in a cycle consistency manner, which is different from existing methods that implicitly achieve controllability through latent diffusion denoising. We also propose a novel and efficient reward strategy that calculates consistency loss by adding noise to input images followed by single-step denoising, thus avoiding the significant computational and memory costs associated with sampling from random Gaussian noise. Experimental results under multiple conditional controls show that ControlNet++ significantly improves controllability without compromising image quality and image-text alignment, offering new insights into controllable visual generation.

\clearpage
\title{ControlNet$++$: Improving Conditional Controls with Efficient Consistency Feedback \\ \textit{Supplementary Material}}


\titlerunning{ControlNet$++$}
\author{
    Ming Li\inst{1} \and
    Taojiannan Yang\inst{1} \and
    Huafeng Kuang\inst{2} \and
    Jie Wu\inst{2} \and \\
    Zhaoning Wang\inst{1} \and
    Xuefeng Xiao\inst{2} \and
    Chen Chen\inst{1}\orcidlink{0000-0003-3957-7061}
}

\authorrunning{Li et al.}

\institute{Center for Research in Computer Vision, University of Central Florida \and
ByteDance}

\maketitle

\vspace{-0.5cm}
\section{Overview of Supplementary}
The supplementary material is organized into the following sections:
\begin{itemize}
    \item Section~\ref{details}: Implementation details for all experiments.
    \item Section~\ref{proof}: Proof for Eq.(7) in the main paper.
    \item Section~\ref{exps}: More experiments and analysis.
    \begin{itemize}
        \item{Section~\ref{subsec:supp_analysis}: Effectiveness of conditioning scale of existing methods.}
        \item{Section~\ref{subsec:human_eval}: Human evaluation on controllability, text guidance and image quaility.}
      \end{itemize}
    \item Section~\ref{impact_and_limitation}: Discussion of broader impact and limitation.
    \item Section~\ref{visualization}: More visualization results.
\end{itemize}

\section{Implementation Details}
\label{details}
\subsection{Dataset Details}
Considering that the training data for ControlNet~\cite{controlnet} has not been publicly released, we need to construct our training dataset.
In this paper, we adhere to the dataset construction principles of ControlNet~\cite{controlnet}, which endeavor to select datasets with more accurate conditional conditions wherever possible. Specifically, for the segmentation condition, previous works have provided datasets with accurately labeled segmentation masks~\cite{ade20k_0,ade20k_1,cocostuff}. Therefore, we opt to train our model using these accurately labeled datasets following ControlNet~\cite{controlnet}. For the Hed, LineArt edge tasks, it is challenging to find datasets with real and accurate annotations. As a result, following ControlNet~\cite{controlnet}, we train the model using the MultiGen20M dataset~\cite{unicontrol}, which is annotated by models, to address this issue. 
Regarding the depth task, existing datasets include masks of certain pixels as having unknown depth values, making them incompatible with the current ControlNet pipeline. Therefore, we also adapt the MultiGen20M depth dataset, which is similar to the dataset constructed by ControlNet~\cite{controlnet}. In terms of the canny edge task, no human labels are required in the process, so we also adapt the MultiGen20M dataset. We provide details of the datasets in Table~\ref{tab:dataset_info}.

\begin{table}[t!]
\caption{Dataset and evaluation details of different conditional controls. $\uparrow$ denotes higher is better, while $\downarrow$ means lower is better.}
\vspace{-0.4cm}
\label{tab:dataset_info}
\resizebox{\textwidth}{!}{%
\begin{tabular}{c|c|c|c|c|c}
\toprule
\textbf{} & \textbf{Segmentation Mask} & \textbf{Canny Edge} & \textbf{Hed Edge} & \textbf{LineArt Edge} & \textbf{Depth Map} \\
\hline
\textbf{Dataset} & ADE20K\cite{ade20k_0,ade20k_1}, COCOStuff\cite{cocostuff} & MultiGen20M\cite{unicontrol} & MultiGen20M\cite{unicontrol} & MultiGen20M\cite{unicontrol} & MultiGen20M\cite{unicontrol} \\
\hline
\textbf{Training Samples} & 20,210 \& 118,287 & 2,560,000 & 2,560,000 & 2,560,000 & 2,560,000 \\
\hline
\textbf{Evaluation Samples} & 2,000 \& 5,000 & 5,000 & 5,000 & 5,000 & 5,000 \\
\hline
\textbf{Evaluation Metric} & mIoU $\uparrow$ & F1 Score $\uparrow$ & SSIM $\uparrow$ & SSIM $\uparrow$ & RMSE $\downarrow$ \\
\bottomrule
\end{tabular}%
}
\end{table}

\begin{table}[t!]
\caption{Details of the reward model, evaluation model, and training loss under different conditional controls. ControlNet* denotes we use the same model to extract conditions as ControlNet~\cite{controlnet}}.
\vspace{-0.4cm}
\label{tab:rm_details}
\resizebox{\textwidth}{!}{%
\begin{tabular}{c|c|c|c|c|c}
\toprule
\textbf{} & \textbf{Seg. Mask} & \textbf{Depth Edge} & \textbf{Canny Edge} & \textbf{Hed Edge} & \textbf{LineArt Edge} \\
\hline
\textbf{Reward Model (RM)} & UperNet-R50 & DPT-Hybrid & Kornia Canny & ControlNet* & ControlNet* \\
\hline
\textbf{RM Performance} & ADE20K(mIoU): 42.05 & NYU(AbsRel): 8.69 & - & - & - \\
\hline
\textbf{Evaluation Model (EM)} & Mask2Former & DPT-Large & Kornia Canny & ControlNet* & ControlNet* \\
\hline
\textbf{EM Performance} & ADE20K(mIoU): 56.01 & NYU(AbsRel): 8.32 & - & - & - \\
\hline
\textbf{Consistency Loss} & CrossEntropy Loss & MSE Loss & MSE Loss & MSE Loss & MSE Loss \\
\hline
\textbf{Loss Weight $\lambda$} & 0.5 & 0.5 & 1.0 & 1.0 & 10 \\
\bottomrule
\end{tabular}%
}
\end{table}

\subsection{Reward Model and Evaluation Details}
In general, we deliberately choose slightly weaker models as the reward model and opt for stronger models for evaluation.
This practice not only ensures the fairness of the evaluation but also helps to determine whether performance improvements result from alignment with the reward model's preferences or from a genuine enhancement in controllability. 
While such an approach is feasible for some tasks (Segmentation, Depth), it becomes challenging to implement for others (Hed, Canny, LineArt Edge) due to the difficulty in finding two distinct reward models. In such cases, we use the same model as both the reward model and the evaluation model. 
We utilize standard evaluation schemes from their respective research fields to evaluate the input conditions and extracted conditions from the generated images, as demonstrated in Section 4.1 of the main paper. We use the same Hed edge detection model and LineArt edge detection model as ControlNet~\cite{controlnet}. We provide details of reward models and evaluation in Table~\ref{tab:rm_details}.

\subsection{Training Details}
The loss weight $\lambda$ for reward consistency loss is different for each condition. Specifically, $\lambda$ is 0.5, 0.5, 1.0, 1.0, and 10 for segmentation mask, depth, hed edge, canny edge, and LineArt edge condition, respectively. For all experiments, we first fine-tune the pre-trained ControlNet until convergence using a batch size of 256 and a learning rate of 1e-5. We then employ the same batch size and learning rate for 10k iterations reward fine-tuning. To this end, the valid training samples for reward fine-tuning is $256 \times 10,000 = 2,560,000$. We set threshold $t_{\text{thre}}=200$ of Eq.8 in the main paper for all experiments.  Diverging from existing methods that use OpenCV's~\cite{opencv} implementation of the canny algorithm, we have adopted Kornia's~\cite{kornia} implementation to make it differentiable. Our codebase is based on the implementation in HuggingFace's Diffusers~\cite{diffusers}, and we do not use classifier-free guidance during the reward fine-tuning process following diffusers.

\begin{figure*}[t!]\centering
    \includegraphics[width=1.0\linewidth]{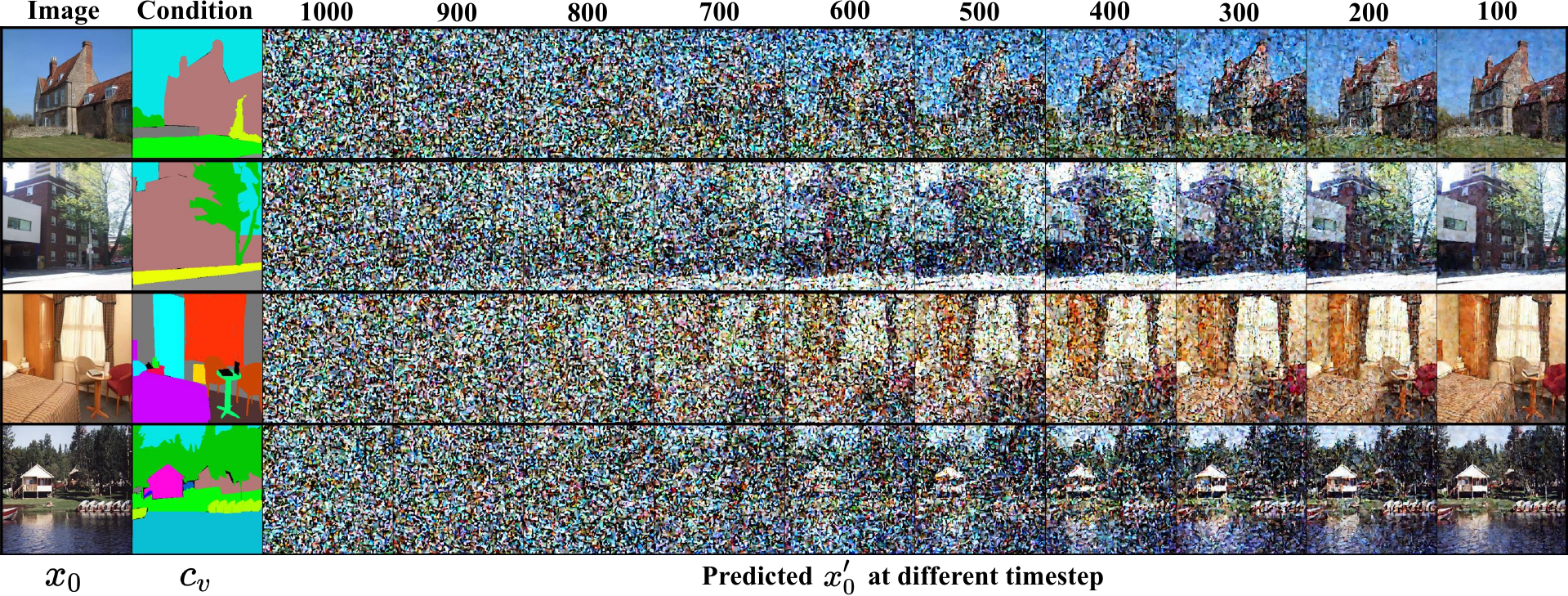}
    \vspace{-0.4cm}
    \caption{
        Illustration of predicted image $x'_0$ at different timesteps $t$. A small timestep $t$ (\emph{i.e.}, small noise $\epsilon_t$) leads to more precise estimation $x'_0 \approx x_0$.
    }
    \label{fig:predicted_x'_0}
\end{figure*}

\section{Proof of Equation 7 in the Main Paper}
\label{proof}
The diffusion models define a Markovian chain of diffusion forward process $q(x_t | x_0)$ by gradually adding noise to input data $x_0$:
\begin{equation}
x_t=\sqrt{\bar{\alpha}_t} x_0+\sqrt{1-\bar{\alpha}_t} \epsilon, \quad \epsilon \sim \mathcal{N}(\mathbf{0}, I),
\label{eq:diffusion_forward}
\end{equation}
at any timestep $t$ we can use the predicted $\epsilon(x'_t, c_v, c_t, t-1)$ to estimate the real noise $\epsilon$ in Eq.~\ref{eq:diffusion_forward}, and the above equation can be transformed through straightforward algebraic manipulation to the following form:
\begin{equation}
\begin{aligned}
& x_t \approx \sqrt{\bar{\alpha}_t} x_0+\sqrt{1-\bar{\alpha}_t} \epsilon_\theta\left(x_t^{\prime}, c_v, c_t, t-1\right),
\\
& x_0 \approx x_0^{\prime}=\frac{x_t^{\prime}-\sqrt{1-\alpha_t} \epsilon_\theta\left(x_t^{\prime}, c_v, c_t, t-1\right)}{\sqrt{\alpha_t}}.
\end{aligned}
\end{equation}
To this end, we can obtain the predicted original image $x'_0$ at any denoising timestep $t$ and use it as the input for reward consistency loss. However, previous work demonstrates that this approximation only yields a smaller error when the time step $t$ is relatively small~\cite{ddpm}. Here we find similar results as shown in Figure~\ref{fig:predicted_x'_0}, which illustrates the predicted $x'_0$ is significantly different at different timesteps. We kindly encourage readers to refer to Section 4.3 and Figure 5 in the DDPM~\cite{ddpm} paper for more experimental results.

\section{More Experiments}
\label{exps}
In this section, we provide additional supplements to the experiments discussed in the main paper, including human evaluation on generated data samples on the Segmentation Mask condition in Sec.~\ref{subsec:human_eval},  analysis on conditioning scale of existing methods such as ControlNet~\cite{controlnet} and T2I-Adapter~\cite{t2i_adapter} in Sec.~\ref{subsec:supp_analysis}.

\subsection{Effectiveness of Conditioning Scale}
\label{subsec:supp_analysis}
\begin{figure*}[h]\centering
    \vspace{-1.0cm}
    \includegraphics[width=1.0\linewidth]{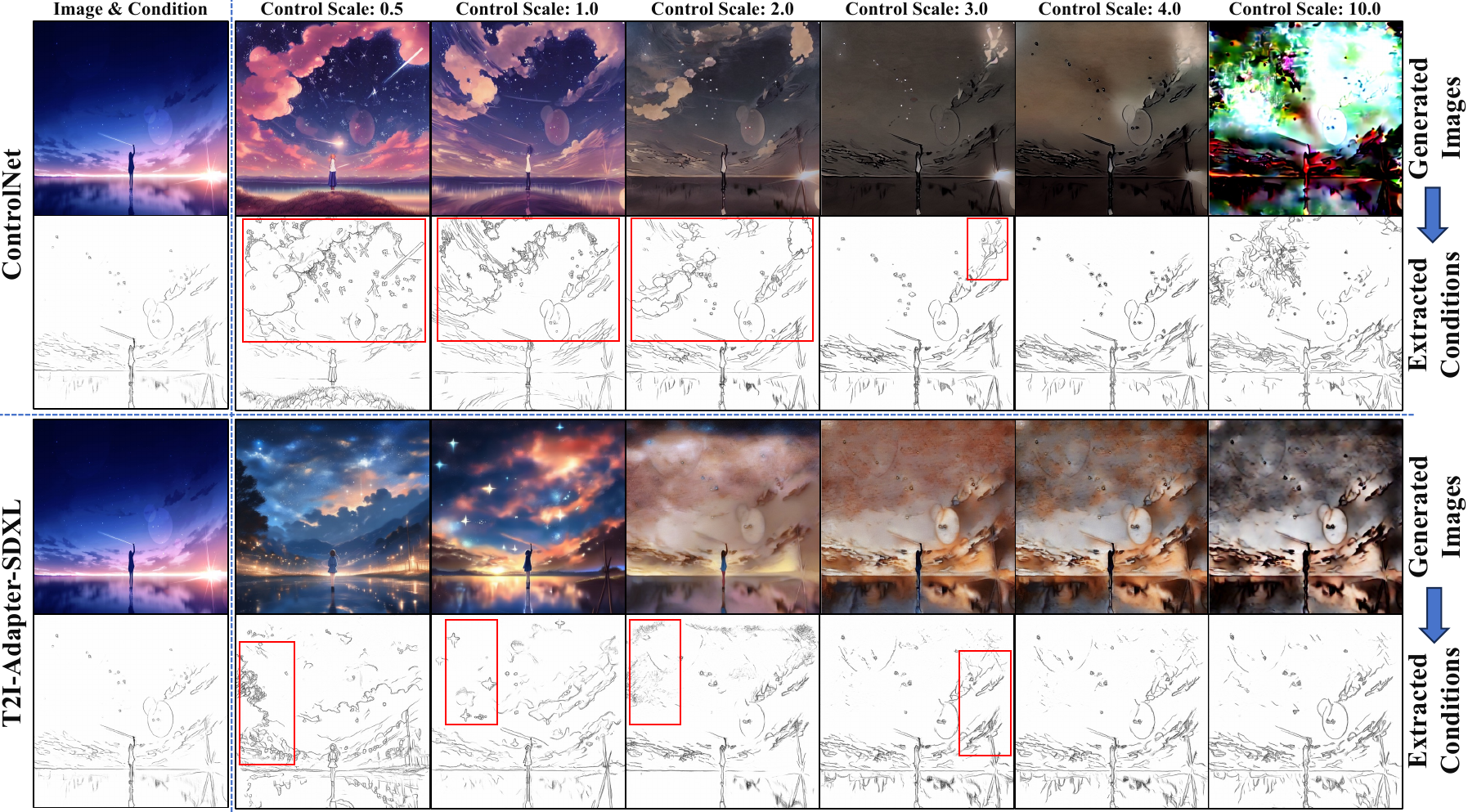}
    \vspace{-0.7cm}
    \caption{
        Naively increasing the weight of image condition embedding compared to text condition embedding in exiting methods (\emph{i.e.}, ControlNet and T2I-Adapter) \textbf{cannot} improve controllability while ensuring image quality. The red boxes in the figures highlight areas where the generated image is inconsistent with the input conditions. Please note that we employ the same line detection model to extract conditions from images.
    }
    \label{fig:conditioning_scale}
\end{figure*}

To simultaneously achieve control based on text prompts and image conditions, existing controllable generation methods perform an addition operation between the image condition features and the text embedding features. The strength of different conditions can be adjusted through a weight value. Hence, an obvious question arises: can better controllability be achieved by increasing the weight of the image condition features? To answer this question, we conduct experiments under different control scales (The weight of image condition feature) in Figure~\ref{fig:conditioning_scale}. It demonstrates that naively increasing the control ratio of image conditions does not enhance controllability and may lead to severe image distortion. 

\subsection{Human Evaluation}
\label{subsec:human_eval}
Following ControlNet, we use a single condition for human evaluation. We ask 20 users (12 in ControlNet paper) to select the best image based on three distinct criteria as shown in Table~\ref{tab:user_study}. Our ControlNet++ offers better controllability without sacrificing image quality or text guidance.

\begin{table}[h]
\centering
\caption{Win rate on ADE20K validation dataset (Segmentation).}
\vspace{-0.1cm}
\label{tab:user_study}
\resizebox{0.8\linewidth}{!}{%
\begin{tabular}{c|c|c|c|c}
\hline
\textit{20 annotators in total} & Ours & ControlNet & T2I-Adapter & UniControl \\ \hline
Image-Mask Alignment & \textbf{76.8\%} & 16.7\% & 2.0\% & 4.5\% \\ \hline
Image Quality & \textbf{26.1\%} & 25.8 \% & 23.6\% & 24.5 \% \\ \hline
Image-Text Alignment & \textbf{25.3\%} & 25.1\% & 24.9\% & 24.7\% \\ \hline
\end{tabular}
}
\end{table}
\vspace{-0.5cm}


\begin{figure*}[th!]\centering
    \includegraphics[width=0.96\linewidth]{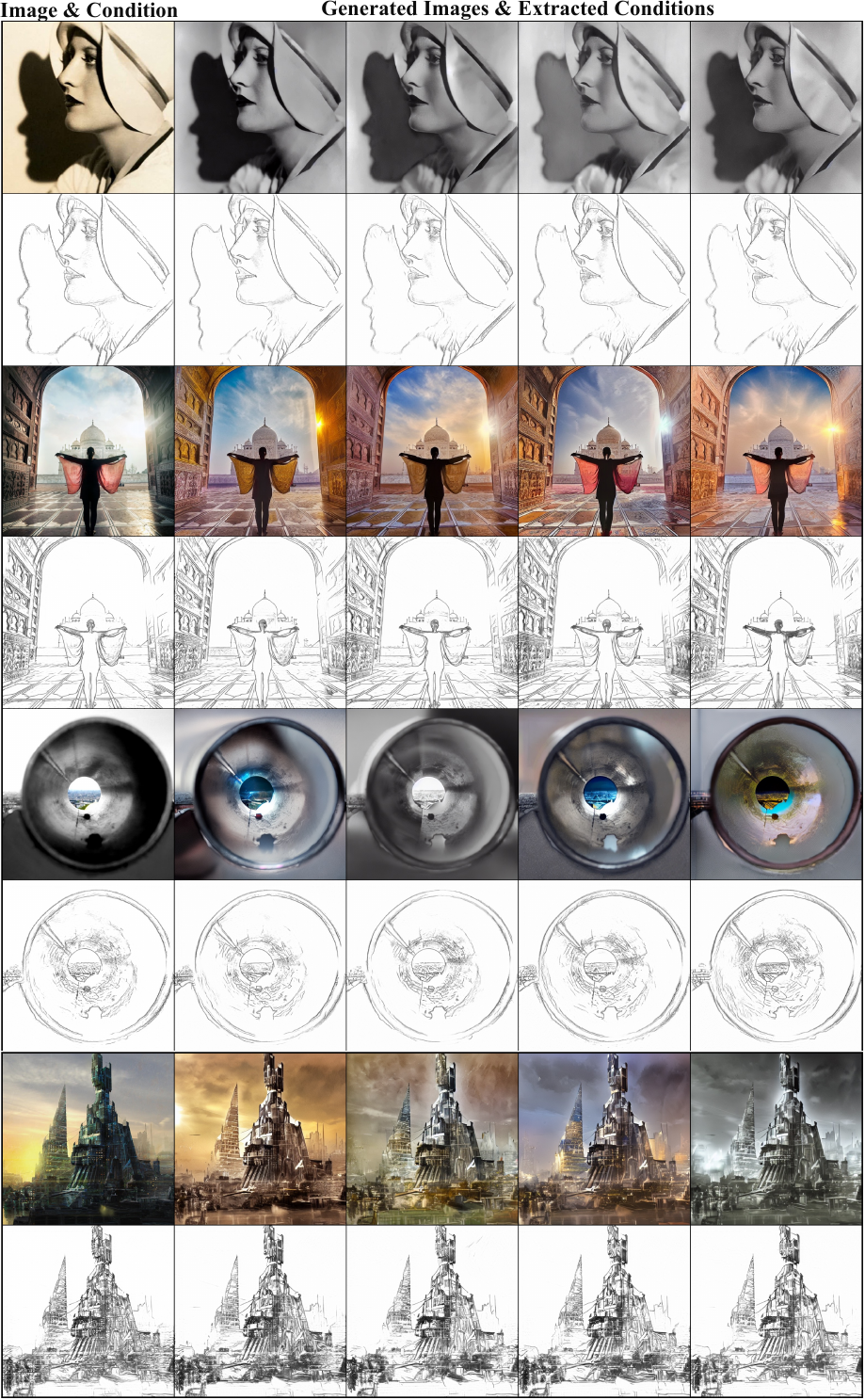}
    \vspace{-0.2cm}
    \caption{
        More visualization results of our ControlNet++ (LineArt Edge)
    }
    \label{fig:vis_lineart}
\end{figure*}

\begin{figure*}[th!]\centering
    \includegraphics[width=0.96\linewidth]{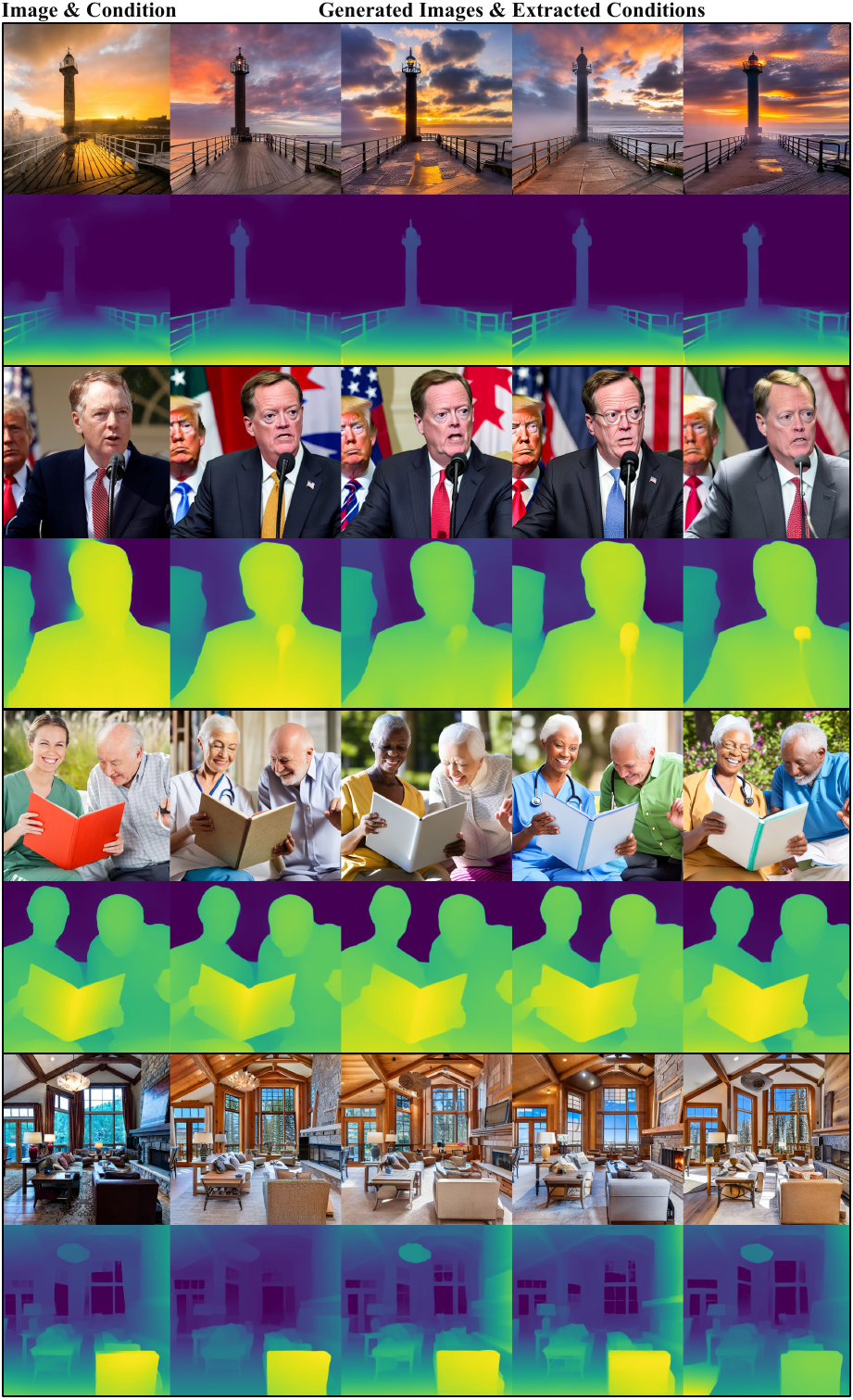}
    \vspace{-0.2cm}
    \caption{
        More visualization results of our ControlNet++ (Depth Map)
    }
    \label{fig:vis_depth}
\end{figure*}

\begin{figure*}[th!]\centering
    \includegraphics[width=0.96\linewidth]{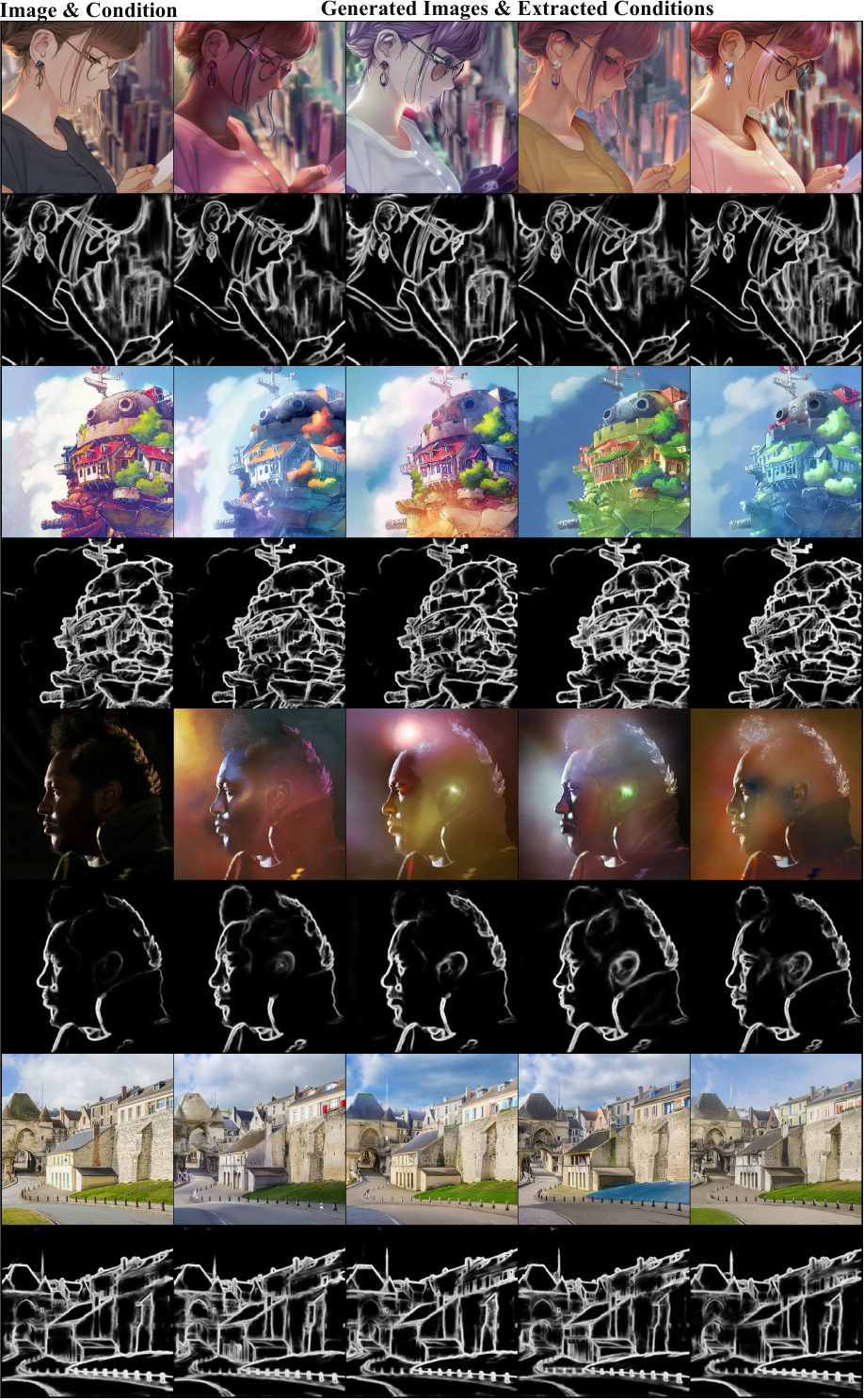}
    \vspace{-0.2cm}
    \caption{
        More visualization results of our ControlNet++ (Hed Edge)
    }
    \label{fig:vis_hed}
\end{figure*}

\begin{figure*}[th!]\centering
    \includegraphics[width=0.96\linewidth]{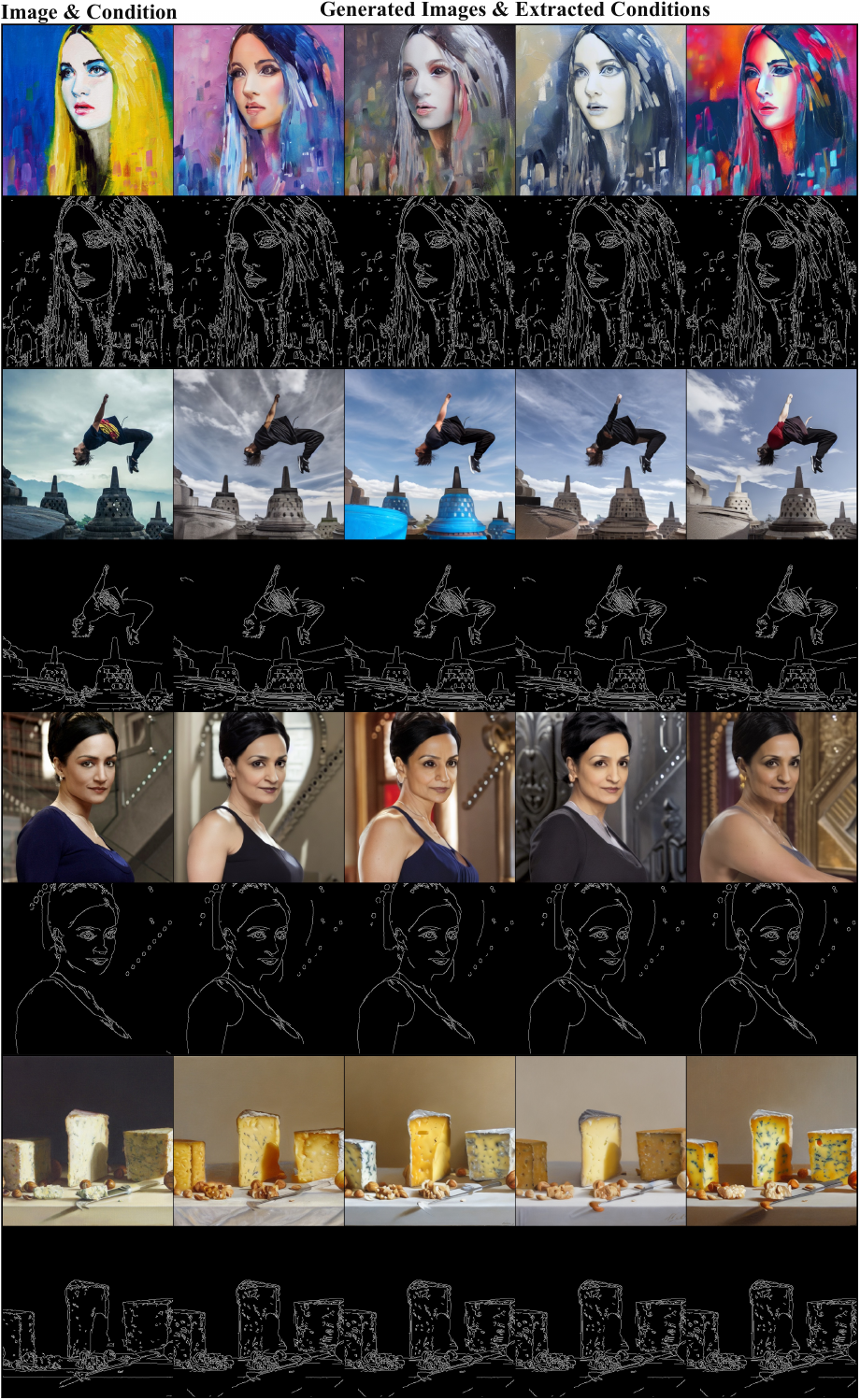}
    \vspace{-0.2cm}
    \caption{
        More visualization results of our ControlNet++ (Canny Edge)
    }
    \label{fig:vis_canny}
\end{figure*}

\begin{figure*}[th!]\centering
    \includegraphics[width=0.96\linewidth]{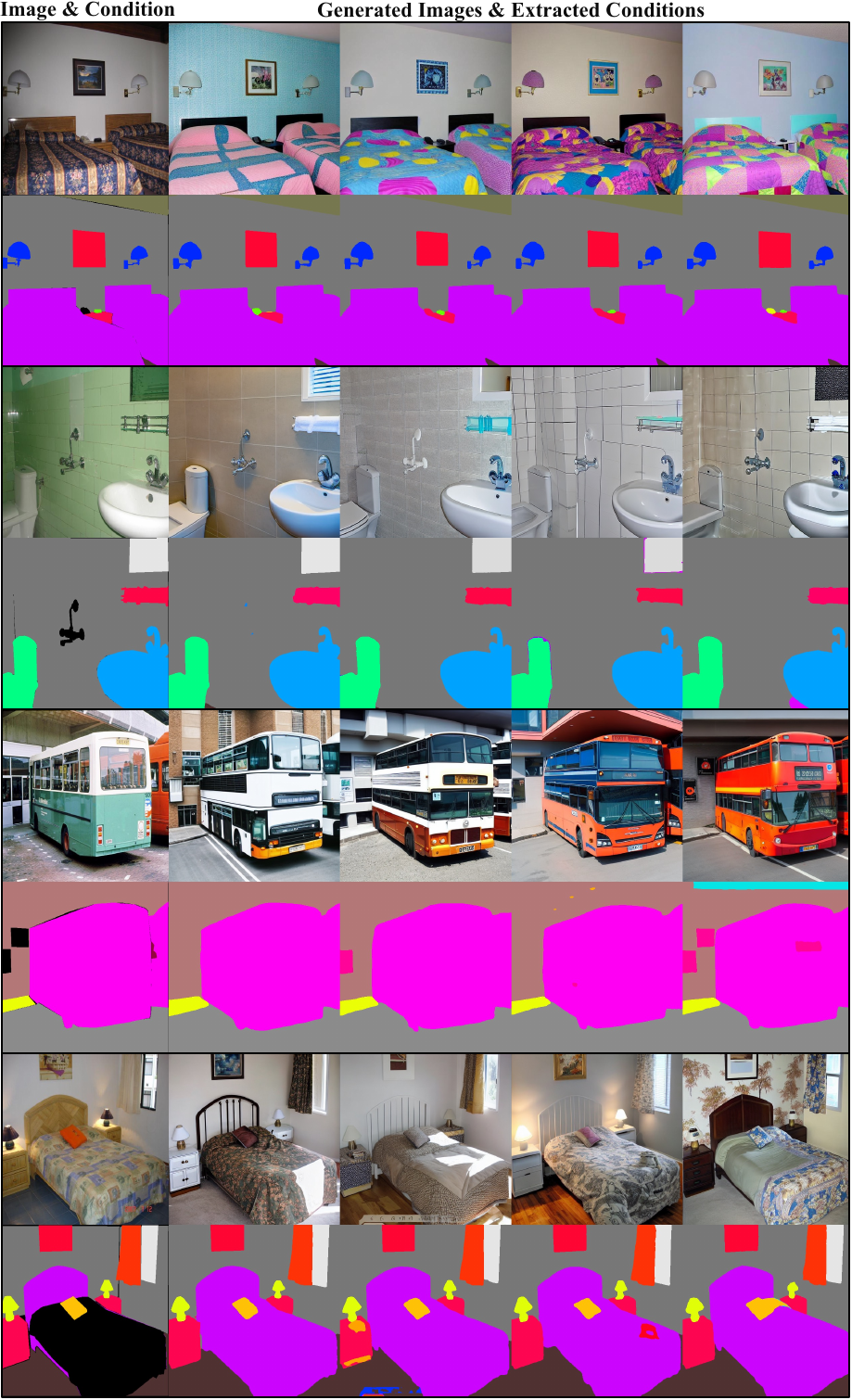}
    \vspace{-0.2cm}
    \caption{
        More visualization results of our ControlNet++ (Segmentation Mask)
    }
    \label{fig:vis_seg}
\end{figure*}

\section{Broader Impact and Limitation}
\label{impact_and_limitation}
In this paper, we use visual discriminative models to evaluate and improve the controllability of text-to-image models. However, we also realize that this work is still insufficient and discuss the following issues:

\noindent\textbf{Conditions Expansion}: While we have achieved notable improvements under six control conditions, our future work aims to broaden the scope by incorporating additional control conditions such as Human Pose and Scribbles. Ultimately, our objective is to control everything.

\noindent\textbf{Beyond Controllability}: While our current focus lies predominantly on controllability, we acknowledge the significance of quality and aesthetic appeal in the generated outputs. To address this, we plan to leverage human feedback to annotate controllability images. Subsequently, we will optimize the controllability model to simultaneously enhance both controllability and aesthetics.

\noindent\textbf{Joint Optimization}: To further enhance the overall performance, we intend to employ a larger set of controllable images for joint optimization of the control network and reward model. This holistic approach would facilitate their co-evolution, leading to further improvements in the final generated outputs. Through our research, we aspire to provide insightful contributions to controllability in text-to-image diffusion models. We hope that our work inspires and encourages more researchers to delve into this fascinating area.

\noindent\textbf{Discussion on the necessity of controllability}: Controllability is important since it allows users to modify image conditions to achieve more flexible and accurate generation. Take LineArt Edge as an example: (1) Freely generating in foreground may change the appearance (\textit{e.g.}, adding a beard for women) that we usually do not expect. (2) Freely generating in background will damage some applications (\textit{e.g.}, blur background). (3) Global free generating may destroy the overall artistic effect of the input image, such as lighting, composition, contrast, etc. Furthermore, we show in Fig.5 of the main paper that more controllable diffusion can in return improve the performance of discriminative models. Beyond image generation, the controllable conditional generation can also be combined with ID preserving methods to perform controllable image editing.

\section{More Visualization}
\label{visualization}
More visualization results across different conditional controls for our image generation are shown in Figures~\ref{fig:vis_lineart},\ref{fig:vis_depth},\ref{fig:vis_hed},\ref{fig:vis_canny},\ref{fig:vis_seg}.

\clearpage


%
%
\bibliographystyle{splncs04}
\bibliography{egbib}

\clearpage  

%
%
\bibliographystyle{splncs04}
\bibliography{egbib}
\end{document}